\documentclass[10pt,twocolumn,letterpaper]{article}
\usepackage{cvpr}
\usepackage{graphicx}
\usepackage{amsmath}
\usepackage{amssymb}
\usepackage{booktabs}
\usepackage{xcolor}
\usepackage{symbols}
\usepackage{multirow}
\usepackage{bbm}
\usepackage{multicol,epigraph,wrapfig}
\usepackage{algorithm, algorithmic}

\newcommand\blfootnote[1]{%
  \begingroup
  \renewcommand\thefootnote{}\footnote{#1}%
  \addtocounter{footnote}{-1}%
  \endgroup
}
\usepackage[pagebackref,breaklinks,colorlinks]{hyperref}
\usepackage[capitalize]{cleveref}
\crefname{section}{Sec.}{Secs.}
\Crefname{section}{Section}{Sections}
\Crefname{table}{Table}{Tables}
\crefname{table}{Tab.}{Tabs.}

\def\cvprPaperID{8286}
\def\confName{CVPR}
\def\confYear{2023}

\begin{document}
\title{Affordances from Human Videos as a Versatile Representation for Robotics}
\author{Shikhar Bahl$^{\star 1,2}$ $\quad$ Russell Mendonca$^{\star 1}$ $\quad$ Lili Chen$^{1}$ $\quad$ Unnat Jain$^{1,2}$ $\quad$ Deepak Pathak$^{1}$ \\ 
\\$^{1}$CMU $\quad$ $^{2}$Meta AI
}

\makeatletter
\let\@oldmaketitle\@maketitle% Store \@maketitle
\renewcommand{\@maketitle}{\@oldmaketitle% Update \@maketitle to insert...
\includegraphics[width=\linewidth]{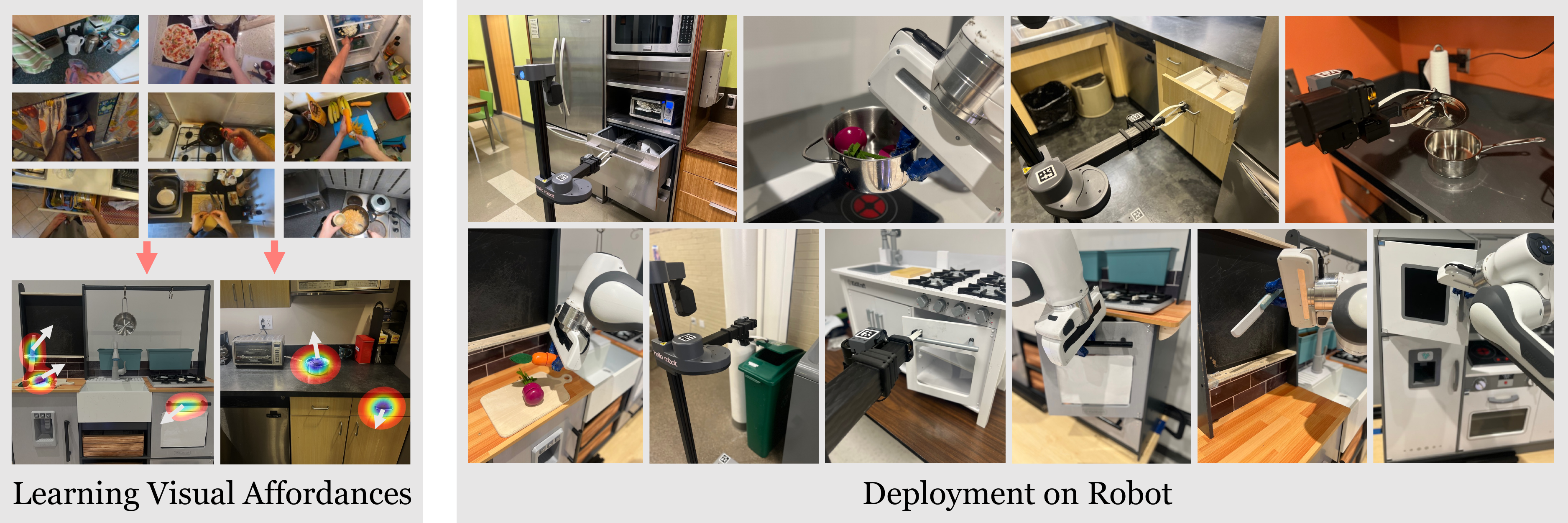}
  \centering
  \captionof{figure}{\small We leverage human videos to learn visual affordances that can be deployed on multiple real robot, in the wild, spanning several tasks and learning paradigms. Videos available at \url{https://robo-affordances.github.io/}. }
  \label{fig:teaser}
  \bigskip}
\makeatother
\maketitle

\begin{abstract}
\vspace{-.5in}

Building a robot that can understand and learn to interact by watching humans has inspired several vision problems. However, despite some successful results on static datasets, it remains unclear how current models can be used on a robot directly. In this paper, we aim to bridge this gap by leveraging videos of human interactions in an environment centric manner. Utilizing internet videos of human behavior, we train a visual affordance model that estimates \emph{where} and \emph{how} in the scene a human is likely to interact. The structure of these behavioral affordances directly enables the robot to perform many complex tasks. We show how to seamlessly integrate our affordance model with four robot learning paradigms including offline imitation learning, exploration, goal-conditioned learning, and action parameterization for reinforcement learning. We show the efficacy of our approach, which we call VRB, across 4 real world environments, over 10 different tasks, and 2 robotic platforms operating in the wild. 

\end{abstract}

\vspace{-.2in}
\epigraph{\textit{The meaning or value of a thing consists of what it affords... what we perceive when we look at objects are their affordances, not their qualities.}}{\textit{J.J. Gibson (1979)}}
\vspace{-0.3in}

\section{Introduction}
\label{sec:intro}
Imagine standing in a brand-new kitchen. Before taking even a single action, we already have a good understanding of how most objects should be manipulated. This understanding goes beyond semantics as we have a belief of where to hold objects and which direction to move them in, allowing us to interact with it. For instance, the oven is opened by pulling the handle downwards, the tap should be turned sideways, drawers are to be pulled outwards, and light switches are turned on with a flick. While things don't always work as imagined and some exploration might be needed, but humans heavily rely on such visual \textit{affordances} of objects to efficiently perform day-to-day tasks across environments~\cite{gibson1966senses,gibson1979ecological}.
Extracting such actionable knowledge from videos has long inspired the vision community.

\begin{figure*}[t!]
    \centering
    \includegraphics[width=\textwidth]{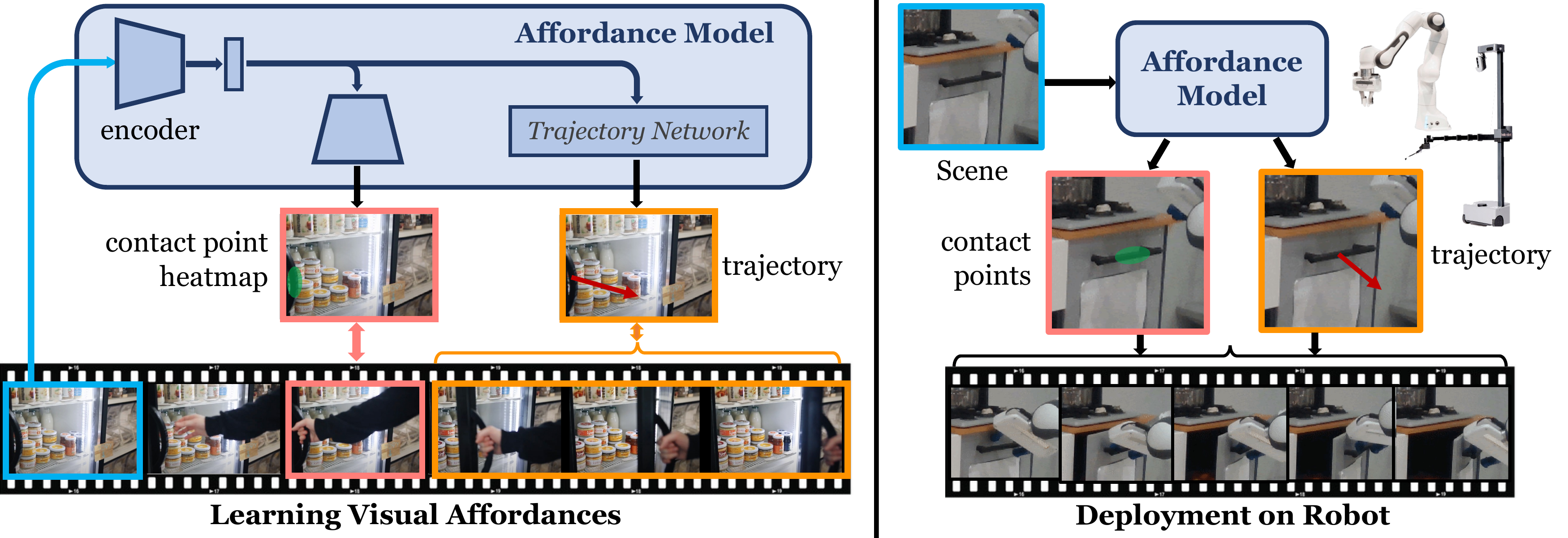} 
     \vspace{-0.22in}
    \caption{\textbf{\ours Overview}. First, we learn an actionable representation of visual affordances from human videos: the model predicts contact points and trajectory waypoints with supervision from future frames. For robot deployment, we query the affordance model and convert its outputs to 3D actions to execute.}
    \vspace{-0.12in}
    \label{fig:method}
\end{figure*}

\blfootnote{$^\star$equal contribution}

More recently, with improving performance on static datasets, the field is increasingly adopting a broader `active' definition of vision through research in egocentric visual understanding and visual affordances from videos of human interaction. With deep learning, methods can now predict heatmaps of where a human would interact~\cite{hotspots, hap} or segmentation of the object being interacted with~\cite{100doh}. Despite being motivated by the goal of enabling downstream robotic tasks, prior methods for affordance learning are tested primarily on human video datasets with no physical robot or in-the-wild experiments. Without integration with a robotic system, even the most basic question of how the affordance should be defined or represented remains unanswered, let alone evaluating its performance.

On the contrary, most robot learning approaches, whether imitation or reinforcement learning, approach a new task or a new environment \textit{tabula rasa}. At best, the visual representation might be pretrained on some dataset~\cite{shah2021rrl,r3m,ma2022vip,Xiao2022,Radosavovic2022,yadav2022offline}. However, visual representations are only a small part of the larger problem. In robotics, especially in continuous control, the state space complexity grows exponentially with actions. Thus, even with perfect perception, knowing what to do is difficult. Given an image, current computer vision approaches can label most of the objects, and even tell us approximately where they are but this is not sufficient for the robot to perform the task. It also needs to know \textit{where} and \text{how} to manipulate the object, and figuring this out from scratch in every new environment is virtually impossible for all but the simplest of tasks. How do we alleviate this clear gap between visual learning and robotics?

In this paper, we propose to rethink visual affordances as a means to bridge vision and robotics. We argue that rich video datasets of humans interacting can offer a lot more actionable information beyond just replacing ImageNet as a pretrained visual encoder for robot learning. Particularly, human interactions are a rich source of how a wide range of objects can be held and what are useful ways to manipulate their state. However, several challenges hinder the smooth integration of vision and robotics. We group them into three parts.
\textit{First}, what is an actionable way to represent affordances?
\textit{Second}, how to learn this representation in a data-driven and scalable manner?
\textit{Third}, how to adapt visual affordances for deployment across robot learning paradigms? 
To answer the first question, we find that \cpoints and \dirs are excellent robot-centric representations of visual affordances, as well as modeling the inherent multi-modality of possible interactions. We make effective use of egocentric datasets in order to tackle the second question. In particular, we reformulate the data to focus on frames without humans for predicting \cpoints and the \dirs. To extract free supervision for this prediction, we utilize off-the-shelf tools for estimating egomotion, human pose, and hand-object interaction. Finally, we show how to seamlessly integrate these affordance priors with different kinds of robot learning paradigms. We thus call our approach \textbf{V}ision-\textbf{R}obotics \textbf{B}ridge (VRB) due to its core goal of bridging vision and robotics.

We evaluate both the quality of our affordances and their usefulness for 4 different robotic paradigms -- imitation and offline learning, exploration, visual goal-reaching, and using the affordance model as a parameterization for action spaces. These are studied via extensive and rigorous real-world experiments on physical robots which span across 10 real-world tasks, 4 environments, and 2 robot hardware platforms. Many of these tasks are performed \textit{in-the-wild} outside of lab environments (see Figure~\ref{fig:teaser}). We find that \ours outperforms other state-of-the-art human hand-object affordance models, and enables high-performance robot learning in the wild without requiring any simulation.
Finally, we also observe that our affordance model learns a good visual representation for robotics as a byproduct.
We highlight that all the evaluations are \textbf{performed in the real world spanning several hundred hours of robot running time} which is a very large-scale evaluation in robotics.

\section{Related Work}
\label{sec:rel}

\noindent\textbf{Affordance and Interaction Learning from Videos.} 
Given a scene, one can predict interactions using geometry-based rules for objects via 3D scene understanding \cite{hassanin2018visual, zhao2013scene, myers2015affordance, myers2014affordance}, estimating 3D physical attributes \cite{eigen2014predicting, bansal2016marr, gupta20113d, zhu2016inferring} or through segmentation models trained on semantic interactions \cite{roy2016multi, sawatzky2017weakly}, and thus require specialized datasets. More general interaction information can be learned from large human datasets ~\cite{li2018eye,Damen2018EPICKITCHENS,Damen2022RESCALING,liu2022hoi4d,VISOR2022,grauman2021ego4d}, to predict object information ~\cite{zhu2014reasoning,furnari2017next} (RGB \& 3D) \cite{bertasius2016first}, 
graphs \cite{dessalene2021forecasting} or environment information \cite{ego-topo,fouhey2012people} such as heatmaps \cite{hap, hotspots}. Approaches also track human poses, especially hands \cite{100doh, liu2020forecasting, rong2021frankmocap,Damen2022RESCALING,hoi,cao2021reconstructing,ye2022hand}. Similarly, in action anticipation and human motion forecasting, high-level semantic or low level actions are predicted using visual history~\cite{koppula2015anticipating, rhinehart2016learning, gao2017red,jain2016recurrent,huang2014action,vondrick2016predicting,abu2018will, Damen2018EPICKITCHENS,
visdial_rl,
visdialq,
brown2020language,furnari2019would,lan2014hierarchical,villegas2017decomposing,grauman2021ego4d, furnari2020rolling,mittal2022learning, mascaro2022intention, girdhar2021anticipative}. Since our observations only have robot arms and no human hands, we adopt a robot-first formulation, only modeling the contact point and post-contact phase of interaction.  

\vspace{0.1in}
\noindent\textbf{Visual Robot Learning.} Learning control from visual inputs directly is an important challenge. Previous works have leveraged spatial structures of convolutional networks to directly output locations for grasping and pushing from just an image of the scene \cite{pinto2015supersizing, zeng2018learning, zeng2020transporter}, which can limit the type of tasks possible. It is also possible to directly learn control end-to-end \cite{levineFDA15, kalashnikov2018qt} which while general, is quite sample inefficient in the real world. It has been common to introduce some form of prior derived from human knowledge, which could take the form of corrective interactions \cite{gutierrez2018incremental, losey2022physical, davchev2020residual}, structured policy spaces \cite{nasiriany2022augmenting, dalal2021accelerating, ratliff2018riemannian, prada2013dmp, amos2019diffmpc, JainWeihs2020CordialSync,bahl2020neural, bahl2020neural, shao2021concept2robot, yan2020learning}, offline robotics data \cite{ebert2021bridge, kumar2021workflow, kumar2020conservative, mandlekar2021matters, rafailov2021offline}, using pretrained visual representations \cite{shah2021rrl, pvr, r3m, mvp,yadav2022offline} or human demonstrations \cite{Sermanet2017TCN, chen2021dvd, bahl2022human, smith2019avid, sharma2019third, shao2021concept2robot}. 

\vspace{0.1in}
\noindent\textbf{Learning Manipulation from Humans.} Extensive work has been done on Learning from Demonstrations (LfD) where human supervision is usually provided through teleoperation (of a joystick or VR interface)~\cite{spranger2018human,zhang2018deep,mohseni2015interactive} or kinesthetic teaching, where a user physically moves the robot arm~\cite{prada2013dmp,calinon2007learning,chu2016learning,elliott2017learning,maeda2017probabilistic}.With both these approaches, collecting demonstrations is tedious and slow. Recently, works have shown alternate ways to provide human demonstrations, via hand pose estimation and retargeting ~\cite{telekinesis,arunachalam2022dexterous,ye2022learning, shaw2022video, qin2022dexmv} in robot hands, but are mostly restricted to tabletop setups. First and third person human demonstrations have been used to train policies directly, transferred either via a handheld gripper \cite{song2020grasping, young2020visual, pari2021surprising} or using online adaptation~\cite{bahl2022human}. In contrast to directly mimicking a demonstration, we learn robot-centric \textit{affordances} from passive human videos that provide a great initialization for downstream robot tasks, unlike previous work which require in-domain demonstrations.

\section{Affordances from Human Videos (\ours)}
\label{sec:app}

\begin{figure*}[t]
    \centering
    \includegraphics[width=\linewidth]{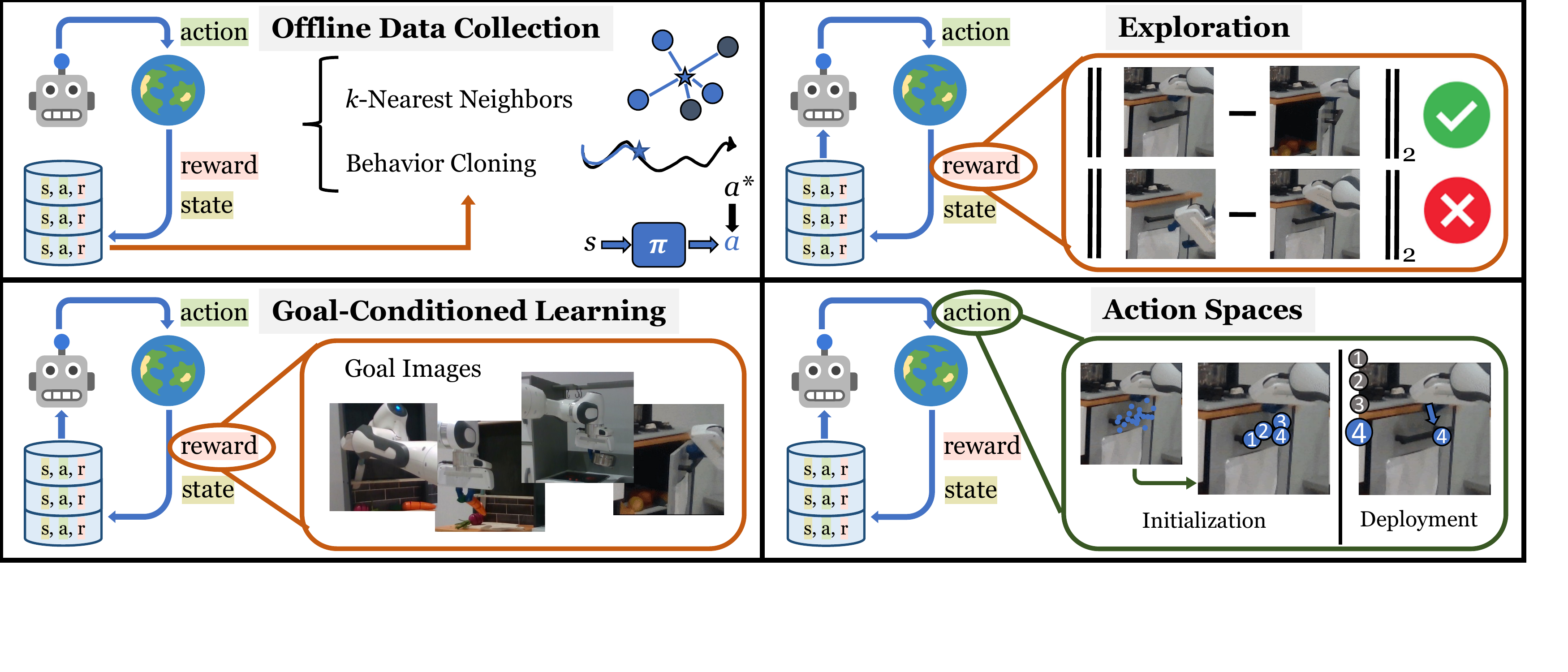}
    \vspace{-0.6in}
    \caption{\textbf{Robot Learning Paradigms} : (a) Offline Data Collection -- Used to investigate the quality of the collected data. (b) Exploration -- The robot needs to use intrinsic rewards to improve (c) Goal-Conditioned Learning -- A desired task is specified via a goal image, used to provide reward. (d) Action Spaces -- Reduced action spaces are easier to search and allow for discrete control.  }
    \label{fig:robot_learning_paradigms}
    \vspace{-0.1in}
\end{figure*}

Our goal is to learn affordance priors from large-scale egocentric videos of human interaction, and then use them to expedite robot learning in the wild. This requires addressing the 
three questions discussed in~\secref{sec:intro} about how to best represent affordances, how to extract them and how to use them across robot learning paradigms.

\subsection{Actionable Representation for Affordances}
\label{sec:method-rep}
Affordances are only meaningful if there is an actor to execute them. For example, a chair has a sitting affordance only if it is possible for some person to sit on it. This property makes it clear that the most natural way to extract human affordances is by watching how people interact with the world. However, what is the right object-centric representation for affordances: is it a heatmap of where the human makes contact? Is it the pre and postcondition of the object? Is it a description of the human interaction? All of these are correct answers and have been studied in prior works \cite{hotspots, hoi, hassanin2018visual}. However, the affordance parameterization should be amenable to deployment on robots.

\begin{figure*}[t!]
    \centering
    \includegraphics[width=\textwidth]{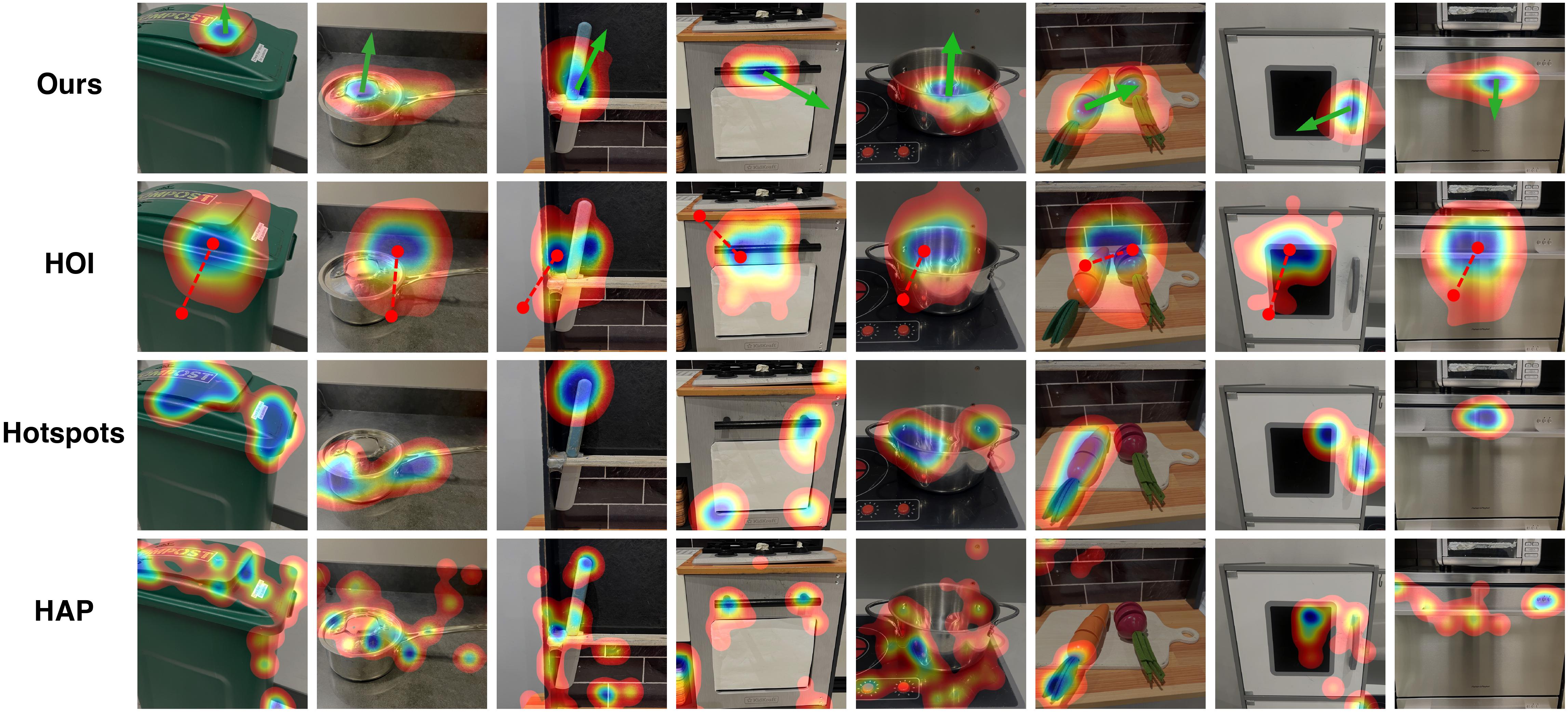}
    \caption{Qualitative affordance model outputs for \texttt{\ours}, \texttt{HOI} \cite{hoi}, \texttt{Hotspots} \cite{hap} and \texttt{HAP} \cite{hap}, showing the predicted contact point region, and post-grasp trajectory (green arrow for \texttt{\ours}, red for \texttt{HOI} \cite{hoi}). We can see that \texttt{\ours} produces the most meaningful affordances.} 
    \label{fig:affordance-comp}
    \vspace{-0.1in}
\end{figure*}

If we want the robot to \textit{a priori} understand how to manipulate a pan (\figref{fig:teaser}, \ref{fig:affordance-comp}) without any interaction, then a seemingly simple solution is to exactly model human movement from videos~\cite{hoi}, but this leads to a human-centric model and will not generalize well because human morphology is starkly different from that of robots. Instead, we take a first-principles approach driven by the needs of robot learning. Knowledge of a robot body is often known, hence reaching a point in the 3D space is feasible using motion planning~\cite{Lavalle2000, lavalle2006planning, karaman2011sampling}. The difficulty is in figuring out where to interact (e.g. the handle of the lid) and then how to move after the contact is made (e.g., move the lid upwards).

Inspired by this, we adopt \cpoints and \dirs as a simple actionable representation of visual affordance that can be easily transferred to robots. We use the notation $c$ for a \cpoint and $\tau$ for \dir, both in the pixel space. Specifically, $\tau = f(I_t, h_t)$, where $I_t$ is the image at timestep $t$, $h_t$ is the human hand location in pixel space, and $f$ is a learned model.
We find that our affordance representation outperforms prior formulations across robots. Notably, the $c$ and $\tau$ abstraction makes the affordance prior agnostic to the morphological differences across robots.

% ======================================================
\subsection{Learning Affordances from Egocentric Videos}
\label{sec:method-extract}
The next question is how to extract $c$ and $\tau$ from human videos in a scalable data-driven manner while dealing with the presence of human body or hand in the visual input. \ours tackles this through a robot-first approach.

\subsubsection{Extracting Affordances from Human Videos}
Consider a video $V$, say of a person opening a door, consisting of $T$ frames \ie $V = \{I_1, ..., I_T\}$. We have a twofold objective --- find \textit{where} and \textit{when} the contact happened, and estimate how the hand moved after contact was made. This is used to supervise the predictive model $f_\theta(I_t)$ that outputs \cpoints and \dirs. To do so, we utilize a widely-adopted hand-object detection model trained on human video data~\cite{100doh}. For each image $I_t$, this produces 2D bounding boxes of the hand $h_t$, and a discrete contact variable $o_t$. Using this information, we filter for frames where $o_t$ indicates a contact in each video, and find the first timestep where contact occurs,  $t_\text{contact}$.  

The pixel-space positions of the hand $\{h_t\}_{t_\text{contact}}^{t'}$ constitute the \dir ($\tau$). To extract contact points $c$, we use the corresponding hand bounding box, and apply skin color segmentation to find all points at the periphery of the hand segment that intersect with the bounding box of the object in contact.  This gives us a set of $N$ contact points $\{c^i\}^N$, where $N$ can differ depending on the image, object, scene and type of interaction. How should the contact points be aggregated to train our affordance model ($f_\theta$)? Some options include predicting the mean of $\{c^i\}^N$, or randomly sampling $c^i$. However, we seek to encourage multi-modality in the predictions, since a scene likely contains multiple possible interactions. To enable this, we fit a Gaussian mixture model (GMM) to the points. Let us define a distribution over contact points to be $p(c)$. We fit the GMM parameters ($\mu_k$, $\Sigma_k$) and weights $\alpha_k$. \begin{equation}
     p(c) = \argmax_{\mu_1, ..., \mu_K, \Sigma_1, ..., \Sigma_K} \sum_{i = 1}^N \sum_{k = 1}^K \alpha_k \mathcal{N}(c^i | \mu_k, \Sigma_k)
\end{equation}

We use these parameters of the above defined GMM with $K$ clusters as targets for $f_\theta$. To summarize, 1) we find the first timestep where contact occurs in the human video, $t_\text{contact}$ 2) For $c$, we fit a GMM to the contact points around the hand at frame $I_{t_\text{contact}}$, parameterized by $\mu_k$, $\Sigma_k$ and 3) we find the post-contact trajectory of the 2D hand bounding box $\{h_t\}_{t_\text{contact}}^{t'}$ for $\tau$.

\vspace{0.05in}
\noindent\textit{Accounting for Camera Motion over Time:}
Consider a person opening a door. Not only do the person's hands move but their body and hence their head also move closer to the handle and then away from it. Therefore, we need to compensate for this egomotion of the human head/camera from time $t_\text{contact}$ to $t'$. We address this by using the homography matrix at timestep $t$, $\mathcal{H}_t$ to project the points back into the coordinates of the starting frame. We obtain the homography matrix by matching features between consecutive frames. We then use this to produce the transformed trajectory $\tau = \mathcal{H}_t\circ\{h_t\}_{t_\text{contact}}^{t'}$.

\vspace{0.05in}
\noindent\textit{Addressing Human-Robot Visual Domain Shift:}
The training videos contain human body or hand in the frame but the human will not be present in downstream robotics task, generating domain shift. We deal with this issue with a simple yet elegant trick: we extract affordances in the frames with humans but then map those affordances back to the first frame when human was yet to enter the scene. For videos in which a human is always in frame, we either crop out the human in the initial frame if there is no interaction yet or discard the frame if the human is always in contact. We compute the \cpoints and \dirs with respect to this human-less frame via the same homography procedure described above. This human-less frame is then used to condition our affordance model.

\subsubsection{Training Affordance Model}
Conditioned on the input image, the affordance model is trained to predict the extracted labels for \cpoints and \dirs. However, naive joint prediction does not work well as the learning problem is inherently multi-modal. For instance, one would pick up a cup differently from a table depending on whether the goal is to pour it into the sink or take a sip from it. We handle this by predicting multiple heatmaps for interaction points using the same model, building a spatial probability distribution.

For ease of notation, we use $(\cdot)_{\theta}$ as a catch-all for all parameterized modules and use $f_\theta$ to denote our complete network. \figref{fig:method} shows an overview of our model. Input image $I_t$ is encoded using a ResNet~\cite{resnet} visual encoder $g^{\text{conv}}_\theta$ to give a spatial latent representation $z_t$, i.e., $g^{\text{conv}}_\theta(I_t) = z_t$. We then project this latent $z_t$ into $K$ probability distributions or heatmaps using deconvolutional layers; concretely, $H_t = g^\text{deconv}_{\theta}(z_t)$. Using a spatial softmax, $\sigma_{\text{2D}}$, we get the estimation of the labels for GMM means, \ie, $\mu_k$. We found that keeping the covariance matrices fixed gave better results. Formally, the loss for \cpoint estimation is:
\begin{equation}
    \mathcal{L}_\text{contact} = \left\Vert\mu_i - \sigma_{\text{2D}}\left(g^\text{deconv}_\theta\left(g^{\text{conv}}_\theta\left(I_t\right)\right)\right)\right\Vert_2
\end{equation}

To estimate \dir, we train a trajectory prediction network, $\mathcal{T}_\theta$, based on the latent representation $z_t$. We find that it is easier to optimize for \textit{relative} shifts, \ie, the direction of movement instead of absolute locations, assuming that the first point $\hat{w}_0$ is 0, since the contact points are already spatially grounded. Based on the success of Transformers for sequential prediction, we employ self-attention blocks~\cite{vaswani2017attention} and train to 
optimize $\mathcal{L}_\text{traj} = \left\Vert\tau - \mathcal{T}_\theta(z_t)\right\Vert_2$. In a given scene, there are many objects a human could interact with, which may or may not be present in the training data. We tackle this uncertainty and avoid spurious correlations by sampling local crops of $I_t$ around the contact points. These serve as the effective input to our network $f_\theta$ and enables better generalization.

\subsection{Robot Learning from Visual Affordances}
\label{sec:method-robot}
Instead of finding a particular way to use our affordance model for robotics, we show that it can bootstrap existing robot learning methods.
In particular, we consider four different robotics paradigms as shown in \figref{fig:robot_learning_paradigms}.

\vspace{-0.08in}
\paragraph{A.$\;\;$ Imitation Learning from Offline Data Collection}
\label{sec:imitation_method}
Imitation learning is conventionally performed on data collected by human demonstrations, teleoperation, or scripted policies -- all of which are expensive and only allow for small-scale data collection\cite{yu2018one, bahl2022human, sharma2019third, argall2009survey, bryson2018applied, levineFDA15}. On the other hand, using the affordance model, $f_\theta(\cdot)$ to guide the robot has a high probability of yielding `interesting' interactions.

Given an image input $I_t$, the affordance model produces $(c, \tau) = f_\theta(I_t)$, and we store $\{(I_t, (c, \tau))\}$ in a dataset $\mathcal{D}$. After sufficient data has been collected, we can use imitation learning to learn control policies, often to complete a specific task. A common approach for task specification is to use \emph{goal images} that show the desired configuration of objects. Given the goal image, the \emph{$k$-Nearest Neighbors} ($k$-NN) approach involves filtering trajectories in $\mathcal{D}$ based on their distance to the goal image in feature space. Further, the top (filtered) trajectories can be used for \emph{behavior cloning} (BC) by training a policy, $\pi(c, \tau| I_t)$. We run both $k$-NN and behavior cloning on datasets collected by different methods in \secref{sec:imitation}. Using the same IL approach for different datasets is also useful for comparing the relative quality of the data. This is because higher relative success for a particular dataset implies that the data is qualitatively better, given that the same IL algorithm achieves worse performance on a different dataset. This indicates that the goal (or similar images) were likely seen during data collection.

\vspace{-0.19in}
\paragraph{B.$\;\;$ Reward-Free Exploration}
The goal of exploration is to discover as many diverse skills as possible which can aid the robot in solving downstream tasks. Exploration methods are usually guided by \textit{intrinsic rewards} that are self-generated by the robotic agent, and are not specific to any task \cite{bellemare2016unifying,
pathakICMl17curiosity, ostrovski2017count,
tang2016exploration,liu2021cooperative,
pong2019skew,TBP,g2p,ramakrishnan2020exploration,mendonca2023alan}. However, starting exploration from scratch is too inefficient in the real world, as the robot can spend an extremely large amount of time trying to explore and still not learn meaningful skills to solve tasks desired by humans. Here our affordance model can be greatly beneficial by bootstrapping the exploration from the predicted affordances allowing the agent to focus on parts of the scene likely to be of interest to humans. 
To operationalize this, we first use the affordance model $f_\theta(.)$ for data-collection. We then rank all the trajectories collected using a task-agnostic exploration metric, and fit a distribution $h$ to the $(c, \tau)$ values of the top trajectories. For subsequent data collection, we sample from $h$ with some probability, and otherwise use the affordance model $f$. This process can then be repeated, and the elite-fitting scheme will bootstrap from highly exploratory trajectories to improve exploration even further. For the exploration metric in our experiments, we maximize \emph{environment change} $\text{EC}( I_i, I_j) =  || \phi(I_{i}) - \phi(I_{j})||_2$, (similar to previous exploration approaches \cite{bahl2022human, parisi2021interesting}) between first and last images in the trajectory, where $\phi$ masks the robot and the loss is only taken on non-masked pixels. 

\vspace{-0.08in}
\paragraph{C.$\;\;$ Goal-Conditioned Learning}
While exploring the environment can lead to interesting skills, consider a robot that already knows its goal. Using this knowledge (\eg an image of the opened door), it supervise its policy search. Goal images are frequently used to specify rewards in RL \cite{wasserman2022lastmile,goyal2022ifor, nair2018visual, pathak2018zero, ghosh2019learning, andrychowicz2017hindsight, nair2017combining,ZhuARXIV2016,mezghan2022memory}. Using our affordance model can expedite the process of solving goal-specified tasks. Similar to the exploration setting, we rank trajectories and fit a distribution $h$ to the ($c, \tau$) values of the top trajectories, but here the metric is to minimize distance to the goal image $I_g$.  The metric used in our experiments is to minimize $\text{EC}(I_T, I_g)$, where $I_T$ is the last image in the trajectory, or to minimize $||\psi(I_g) - \psi(I_T) ||_2^2 $, where $\psi$ is a feature space. Akin to exploration, subsequent data collection involves sampling from $h$ and the affordance model~$f$.

\paragraph{D.$\;\;$ Affordance as an Action Space}
Unlike games with discrete spaces like Chess and Go where reinforcement learning is deployed \textit{tabula rasa}, robots need to operate in continuous action spaces that are difficult to optimize over. A pragmatic alternative to continuous action spaces is parameterizing them in a spatial manner and assigning a primitive (e.g. grasping, pushing or placing) to each location~\cite{zeng2021pushing, zeng2018learning, shridhar2022cliport}. 
While this generally limits the type of tasks that can be performed, our affordance model already seeks out interesting states, due to the data it is trained on. We first query the affordance model on the scene many times to obtain a large number of  predictions. We then fit a GMM to these points to obtain a discrete set of $(c, \tau)$ values, and now the robot just needs to search over this space.

\section{Experimental Setup and Results}
\label{sec:exp}
Through the four robot learning paradigms, shown in \figref{fig:robot_learning_paradigms}, we seek to answer the following questions: (1) Does our model enable a robot to collect \emph{useful data} (imitation from offline data)?, (2) How much benefit does \ours provide to \emph{exploration} methods?, (3) Can our method enable \emph{goal-conditioned} learning?, and (4) Can our model be used to define a structured \emph{action space} for robots?
Finally, we also study whether our model learns meaningful \emph{visual representations} for control as a byproduct and also analyze the \emph{failure modes} and how they differ from prior work.

\vspace{0.06in}\noindent\textbf{{Robotics Setup}}$\quad$ We use two different robot platforms - the Franka Emika Panda arm and the Hello Stretch mobile manipulator. We run the Franka on two distinct play kitchen environments and test on tasks that involve interacting with a cabinet, a knife and some vegetables, and manipulation of a a shelf and a pot. The Hello robot is tested on multiple in-the wild tasks outside lab settings, including opening a garbage can, lifting a lid, opening a door, pulling out a drawer, and opening a dishwasher (\figref{fig:teaser}). We also provide support for a simulation environment on the Franka-Kitchen benchmark \cite{d4rl}. Details can be found in the Appendix. 

\vspace{0.06in}\noindent\textbf{{Observation and Action space}}$\quad$
For each task, we estimate a task-space image-crop using bounding boxes \cite{zhou2022detecting}, and pass random sub-crops to $f_\theta$. The prediction for contact points $c$ and post-contact trajectory $\tau$ is in pixel space, which are projected into 3D for robot control using a calibrated robot-camera system (with an Intel RealSense D415i). The robot operates in 6DOF end-effector space -- samples a rotation, moves to a contact point, grasps, and then moves to a post-contact position (see \secref{sec:method-rep}).

\noindent\textbf{{Baselines and Ablations: }} We compare against prior work that has tried to predict heatmaps from human video : 1) Hotspots~\cite{hotspots} 2) Hands as Probes (\texttt{HAP})~\cite{hap}, a modified version for our robot setup of Liu~\etal~\cite {hoi} that predicts contact region and forecast hand poses: 3) \texttt{HOI}~\cite{hoi} and 4) a baseline that samples affordances at random (\texttt{Random}). HAP and  Hotspots only output a contact point, and we randomly select a post-contact direction. More details are available in the Appendix.

\begin{table}[t!]
\centering
\resizebox{1\linewidth}{!}{%
\Huge
\begin{tabular}{lcccccccc}
\toprule
 & \textbf{Cabinet} & \textbf{Knife} & \textbf{Veg} & \textbf{Shelf}  & \textbf{Pot}  & \textbf{Door} & \textbf{Lid} & \textbf{Drawer} \\
\midrule
\multicolumn{9}{l}{\textit{k-Nearest Neighbors}:}\vspace{0.4em}\\
\texttt{HOI}  & 0.2 & 0.1 & 0.1 & 0.6 & 0.0 & 0.4 & 0.0 & 0.6  \\
\texttt{HAP}  & 0.3 & 0.0 & 0.3 & 0.0 & 0.1 & 0.2 & 0.0 & 0.1  \\
\texttt{Hotspots}  & 0.4 & 0.0 & 0.1 & 0.0 & \textbf{0.5} & 0.4 & 0.3 &  0.5\\
\texttt{Random}  & 0.3 & 0.0 & 0.1 & 0.3 & 0.4 & 0.2 & 0.1 & 0.2 \\
\midrule
\textbf{\texttt{\ours} (ours)}  & \textbf{0.6} & \textbf{0.3} & \textbf{0.6} & \textbf{0.8} & 0.4 & \textbf{1.0} & \textbf{0.4} &  \textbf{1.0} \\
\midrule
\multicolumn{9}{l}{\textit{Behavior Cloning}:}\vspace{0.4em}\\
\texttt{HOI}& 0.3 & 0.0 & 0.3 & 0.0 & 0.1 & 0.2 & 0.0 & 0.1  \\
\texttt{HAP} & 0.5 & 0.0 & \textbf{0.4} & 0.0 & 0.3 & 0.1 & 0.0 &  0.1\\
\texttt{Hotspots}  & 0.2 & 0.0 & 0.0 & 0.0 & \textbf{0.8} & 0.1 & 0.0 &  0.7\\
\texttt{Random}  & 0.1 & \textbf{0.1} & 0.1 & 0.0 & 0.2 & 0.1 & 0.0 & 0.0 \\
\midrule
\textbf{\texttt{\ours} (ours)}  & \textbf{0.6} & \textbf{0.1} & 0.3 & \textbf{0.3} & \textbf{0.8} & \textbf{0.9} & \textbf{0.2} & \textbf{0.9} \\
\bottomrule
\end{tabular}}
\vspace{-0.1in}
\caption{ \textbf{Imitation Learning}: Success rate for $k$-NN and Behavior Cloning on collected offline data using various affordance models. We find that \ours vastly outperforms prior approaches, indicating better quality of data.}
\label{tab:dc_success}
\vspace{-0.2in}
\end{table}
\vspace{-0.05in}
\subsection{Quality of Collected Data for Imitation}
\label{sec:imitation}
\vspace{-0.05in}
We investigate \ours as a tool for useful data collection. We evaluate this on both our robots across 8 different environments, with results in~\tabref{tab:dc_success}. These are all unseen scenarios (not in train set). Tasks are specified for each environment using goal images (eg - open door, lifted pot etc), and we use the data collected (30-150 episodes) for two established offline learning methods: (1) k-Nearest Neighbors ($k$-NN) and (2) Behavior Cloning. $k$-NN \cite{pari2021surprising} finds trajectories in the dataset that are close (via distance in feature space~\cite{r3m}) to the goal image. We run the 10-closest trajectories to the goal image and record whether the robot has achieved the task specified in the goal image. For behavior cloning, we train a network supervised with (image, waypoint) pairs from the collected dataset, and the resulting policy is run 10 times on the real system. With both $k$-NN and BC, our method outperforms prior tasks on 7 out of 8 tasks, with an average success rate of 57 \%, with the runner-up method (Hotspots \cite{hotspots}) only getting 25 \%.  This shows that \ours leads to much better data offline data quality, and thus can lead to better imitation learning performance. We additionally test for grasping held-out \textit{rare} objects such as VR remotes or staplers, and find that \ours outperforms baselines. Details can be found in the Appendix.

\subsection{Reward-Free Exploration}
\label{sec:exploration}
\begin{figure}[h!]
    \centering
    \begin{subfigure}[b]{0.49\linewidth}
    \includegraphics[width=\linewidth]{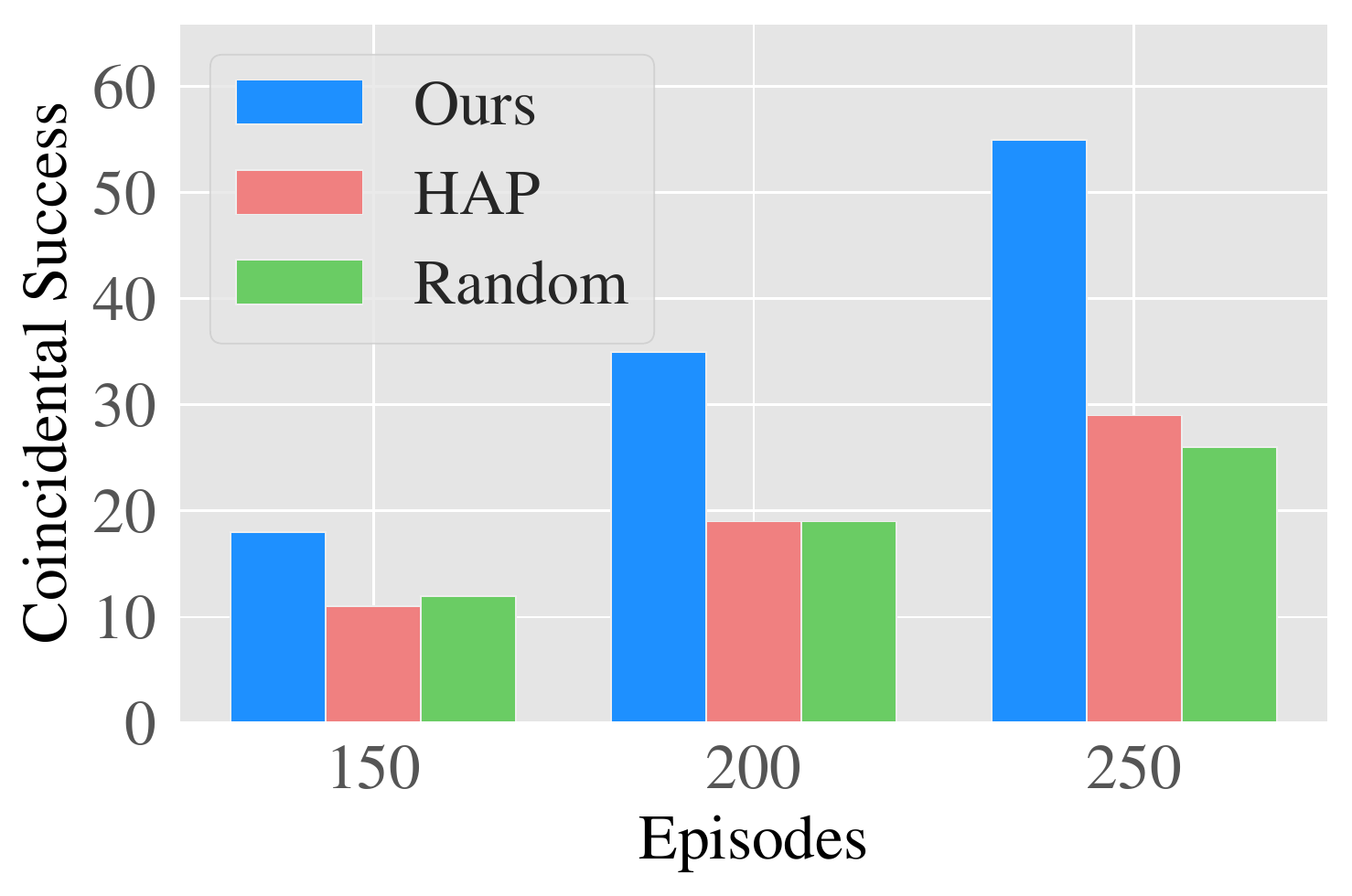}
    \caption{Cabinet}
    \end{subfigure}
    \begin{subfigure}[b]{0.49\linewidth}
    \includegraphics[width=\linewidth]{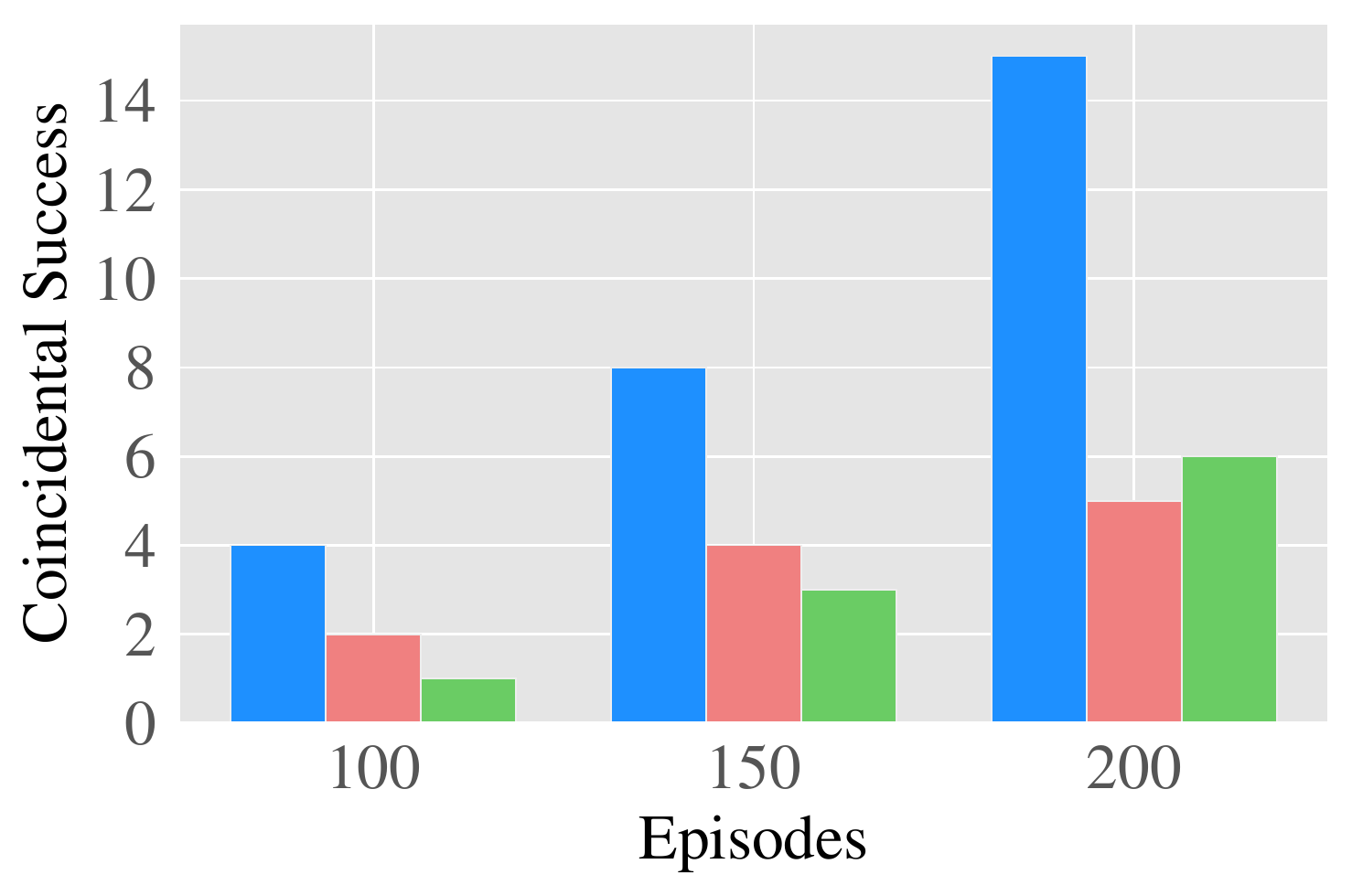}
    \caption{Knife}
    \end{subfigure}
    \begin{subfigure}[b]{0.49\linewidth}
    \includegraphics[width=\linewidth]{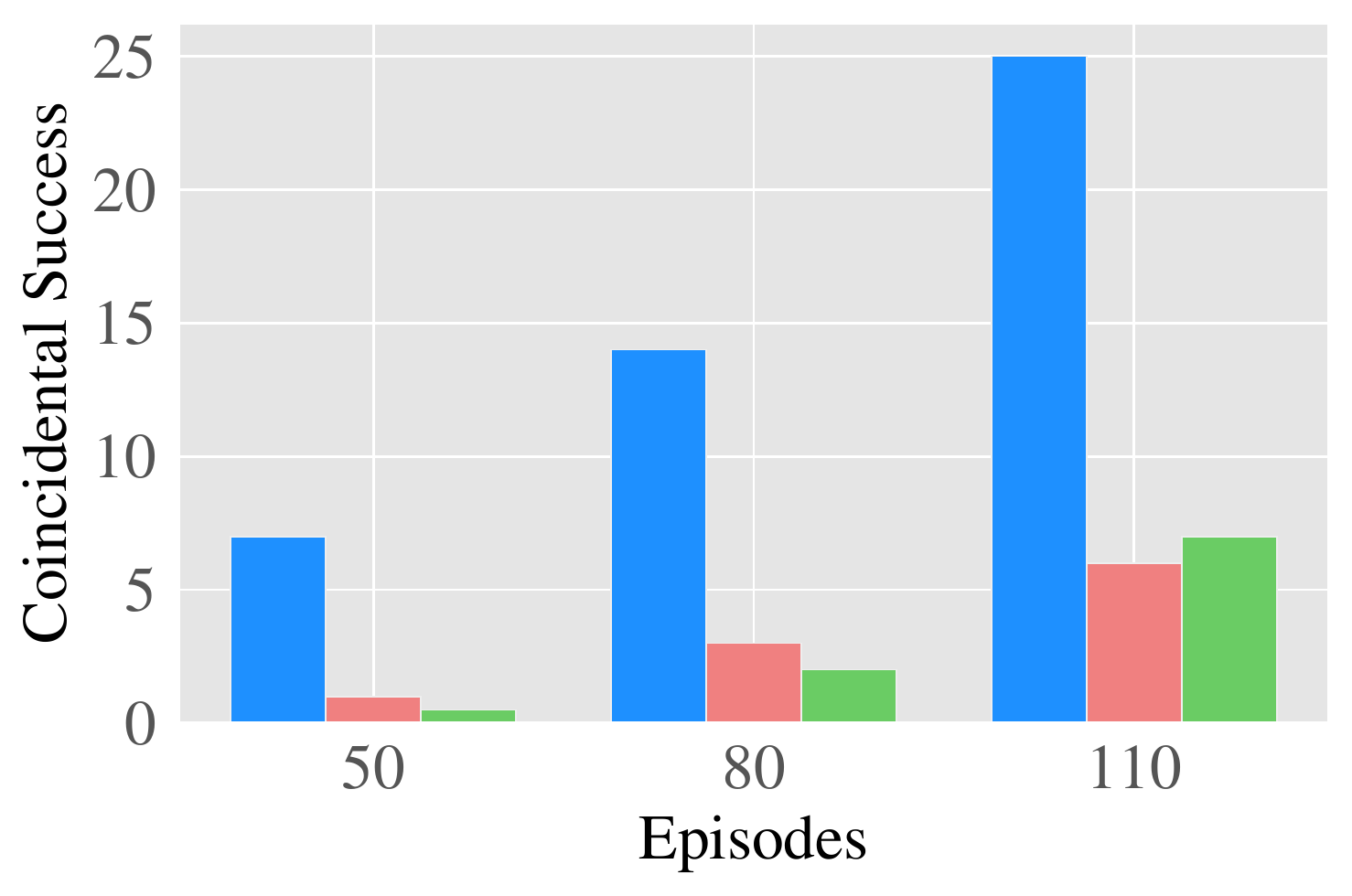}
    \caption{Stovepot}
    \end{subfigure}
    \begin{subfigure}[b]{0.49\linewidth}
    \includegraphics[width=\linewidth]{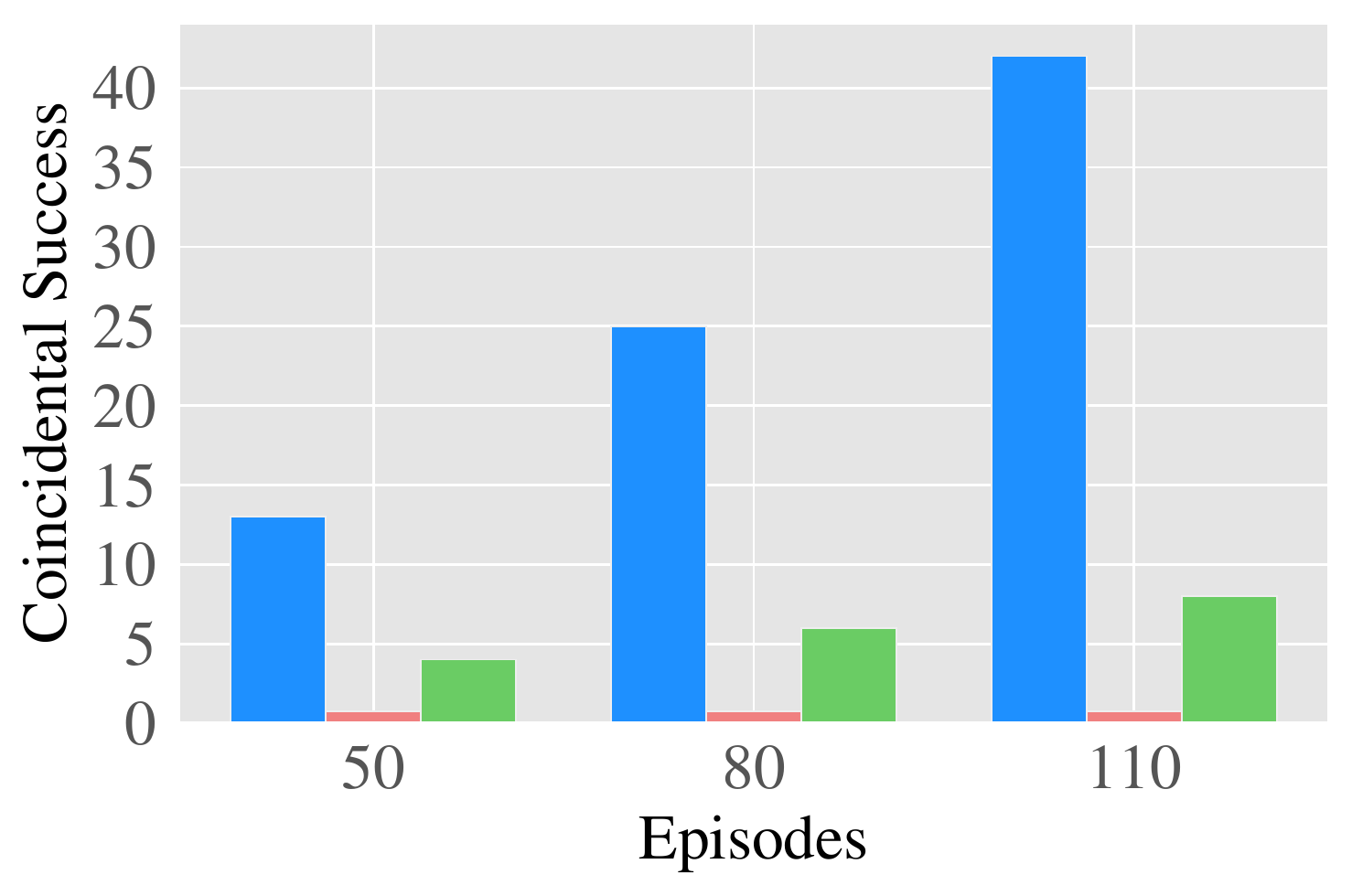}
    \caption{Shelf}
    \end{subfigure}
    \vspace{-0.1in}
    \caption{\textbf{Exploration}: Coincidental success of \ours in comparison to random exploration or the exploration based on HAP~\cite{hap}.}
    \vspace{-0.25in}
    \label{fig:exploration-coinc-plots}
\end{figure}

Here we study self-supervised exploration with no external rewards.
We utilize environment change, \ie, change in the position of objects as a task-agnostic metric for exploration \cite{bahl2022human}. For improved exploration, we bias sampling towards trajectories with a higher environment change metric. 
To evaluate the quality of exploration data, we measure how often does the robot achieves coincidental success \ie reach a goal image configuration without having access to it. As shown in \figref{fig:exploration-coinc-plots}, we obtain consistent improvements over HAP~\cite{hap} and random exploration raising performance multiple fold -- from $3\times$ to $10\times$, for every task. 

\subsection{Goal-Conditioned Learning}
\label{sec:goal}
\begin{figure}[h!]
    \centering
    \begin{subfigure}[b]{0.49\linewidth}
    \includegraphics[width=\linewidth]{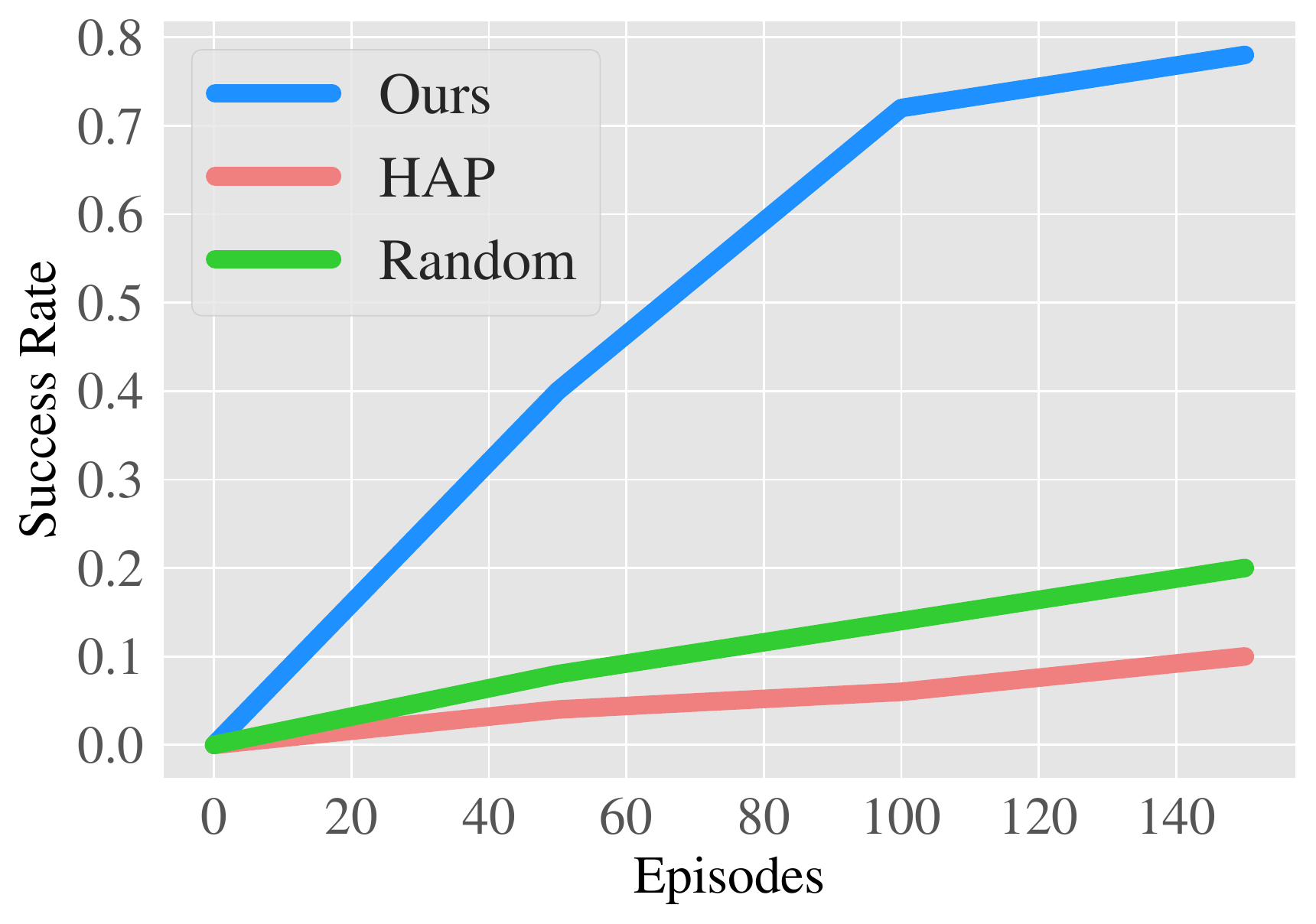}
    \caption{Door}
    \end{subfigure}
    \begin{subfigure}[b]{0.49\linewidth}
    \includegraphics[width=\linewidth]{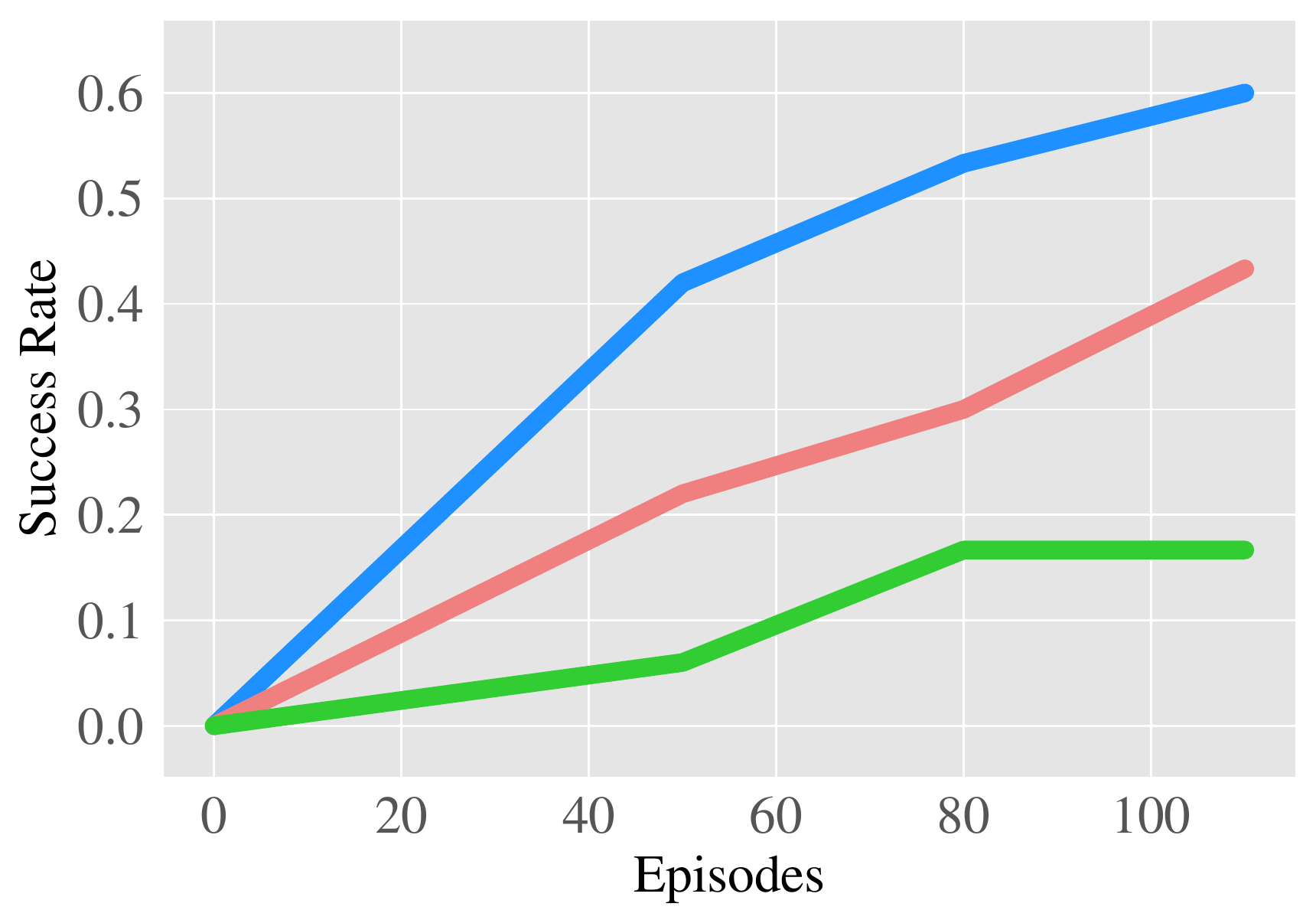}
    \caption{Veggies}
    \end{subfigure}
    \begin{subfigure}[b]{0.49\linewidth}
    \includegraphics[width=\linewidth]{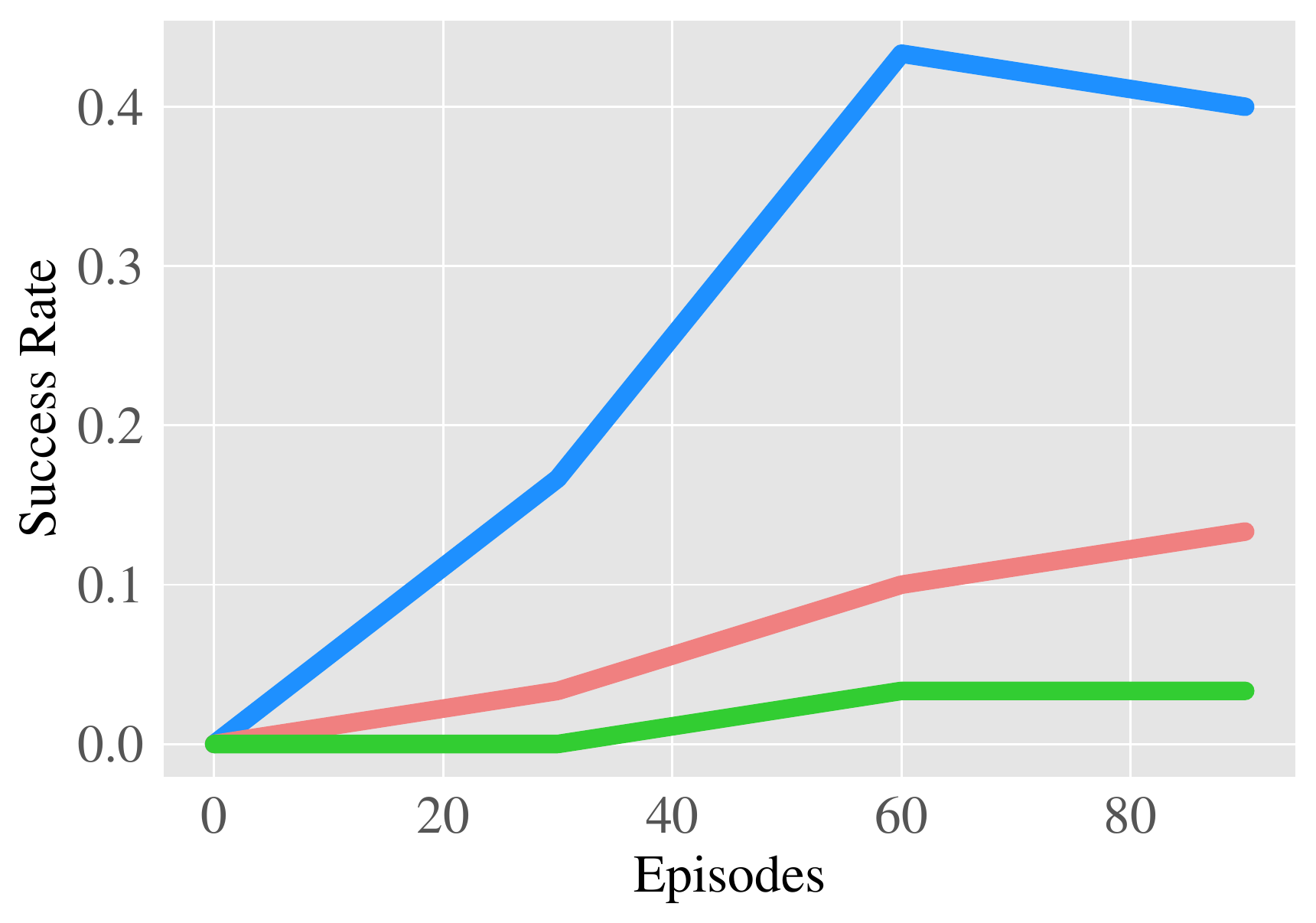}
    \caption{Lid}
    \end{subfigure}
    \begin{subfigure}[b]{0.49\linewidth}
    \includegraphics[width=\linewidth]{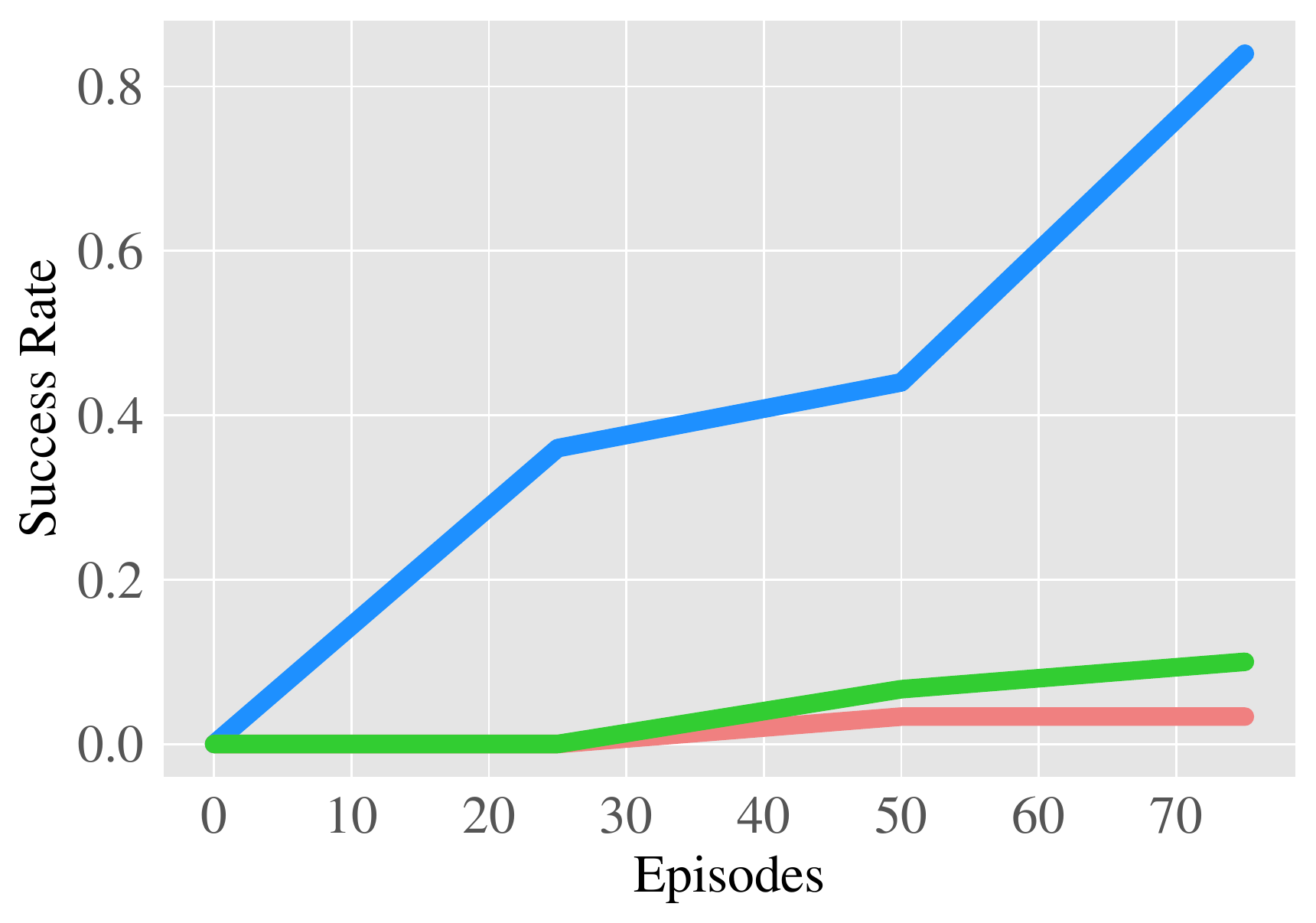}
    \caption{Dishwasher}
    \end{subfigure}
    \begin{subfigure}[b]{0.49\linewidth}
    \includegraphics[width=\linewidth]{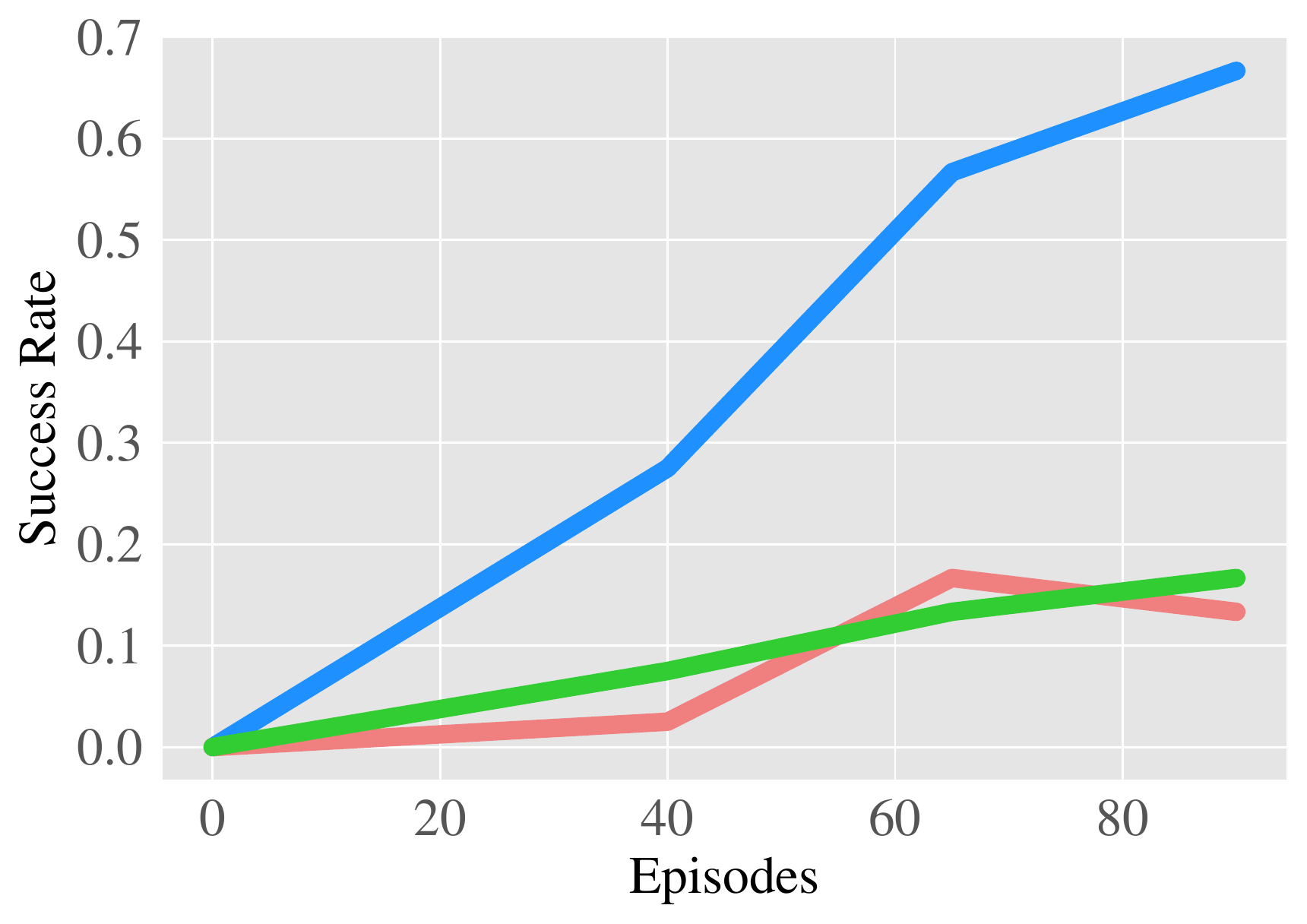}
    \caption{Drawer}
    \end{subfigure}
    \begin{subfigure}[b]{0.49\linewidth}
    \includegraphics[width=\linewidth]{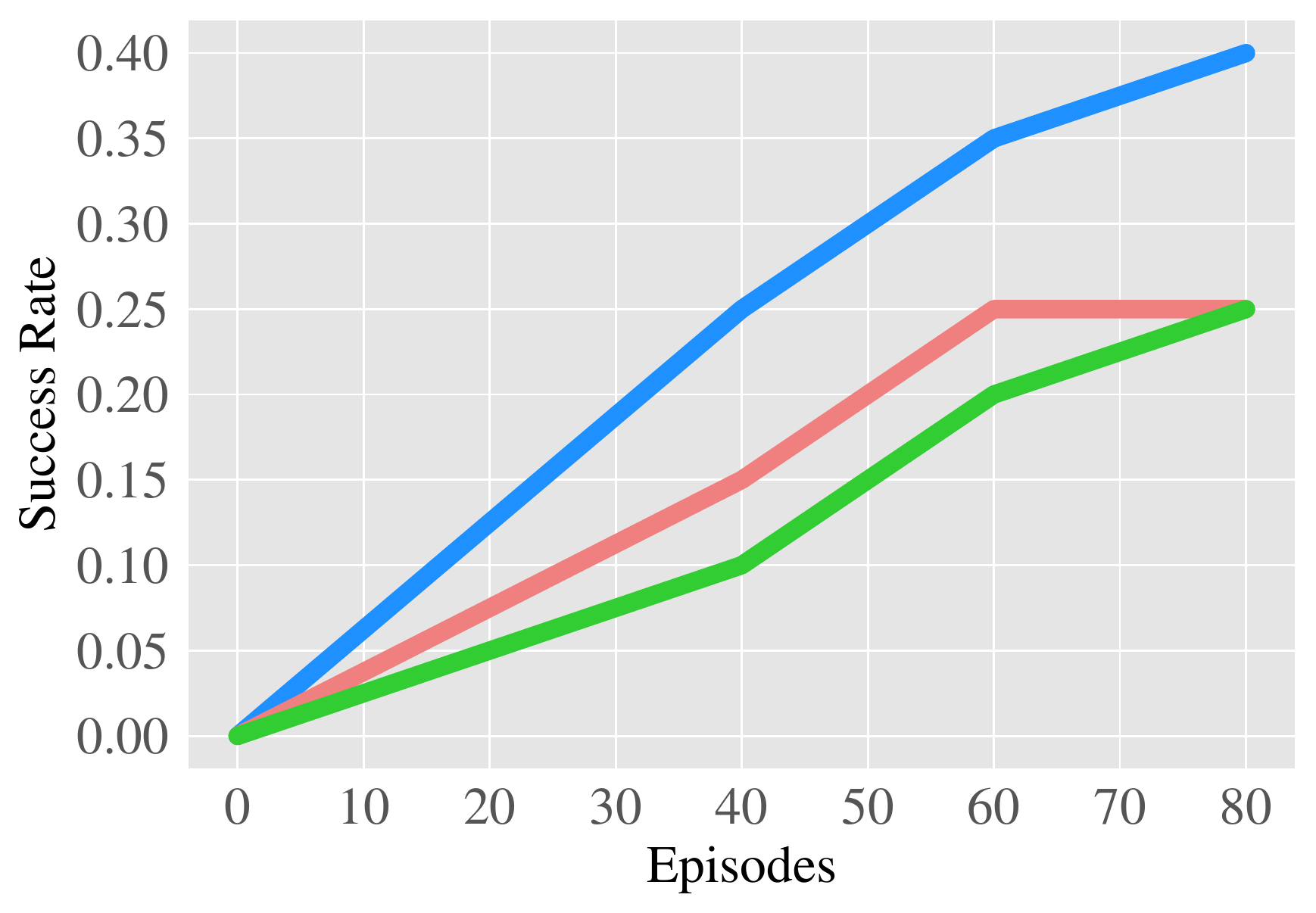}
    \caption{Garbage Can}
    \end{subfigure}
    \vspace{-0.1in}
    \caption{\textbf{Goal-conditioned Learning}:
    Success rate for reaching goal configuration for six different tasks. Sampling via \ours leads to faster learning and better final performance.
}    
    \vspace{-0.15in}
    \label{fig:goal-plots}
\end{figure}

The previous settings help robots improve their behaviors with data without an external reward or goal. 
Here we focus on goal-driven robot learning.
Goals are often specified through images of the goal configuration. 
Note that goal images are also used in~\secref{sec:imitation} but as part of a static dataset to imitate. Here, the robot policy is updated with new data being added to the buffer. 
We sample this dataset for trajectories that minimize visual change with respect to the goal image. As shown in \figref{fig:goal-plots}, \ours learns faster and better HAP~\cite{hap} and Random on this robot learning paradigm, over six diverse tasks.

\subsection{Affordance as an Action Space}
\label{sec:action}
\begin{figure}[t!]
    \centering
    \begin{subfigure}[b]{0.46\linewidth}
    \includegraphics[width=\linewidth]{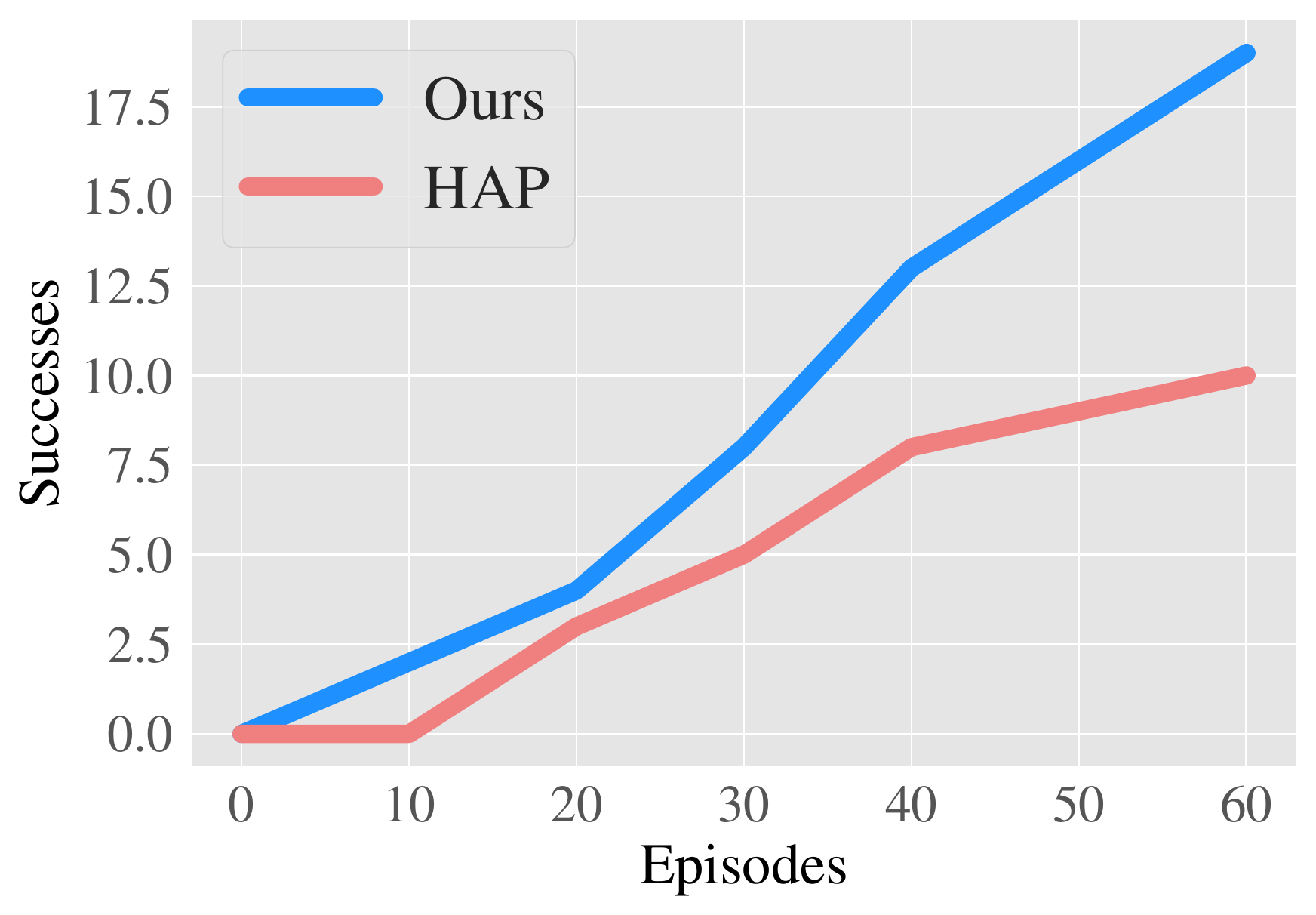}
    \vspace{-0.18in}
    \caption{Cabinet}
    \end{subfigure}
    \quad
    \begin{subfigure}[b]{0.46\linewidth}
    \includegraphics[width=\linewidth]{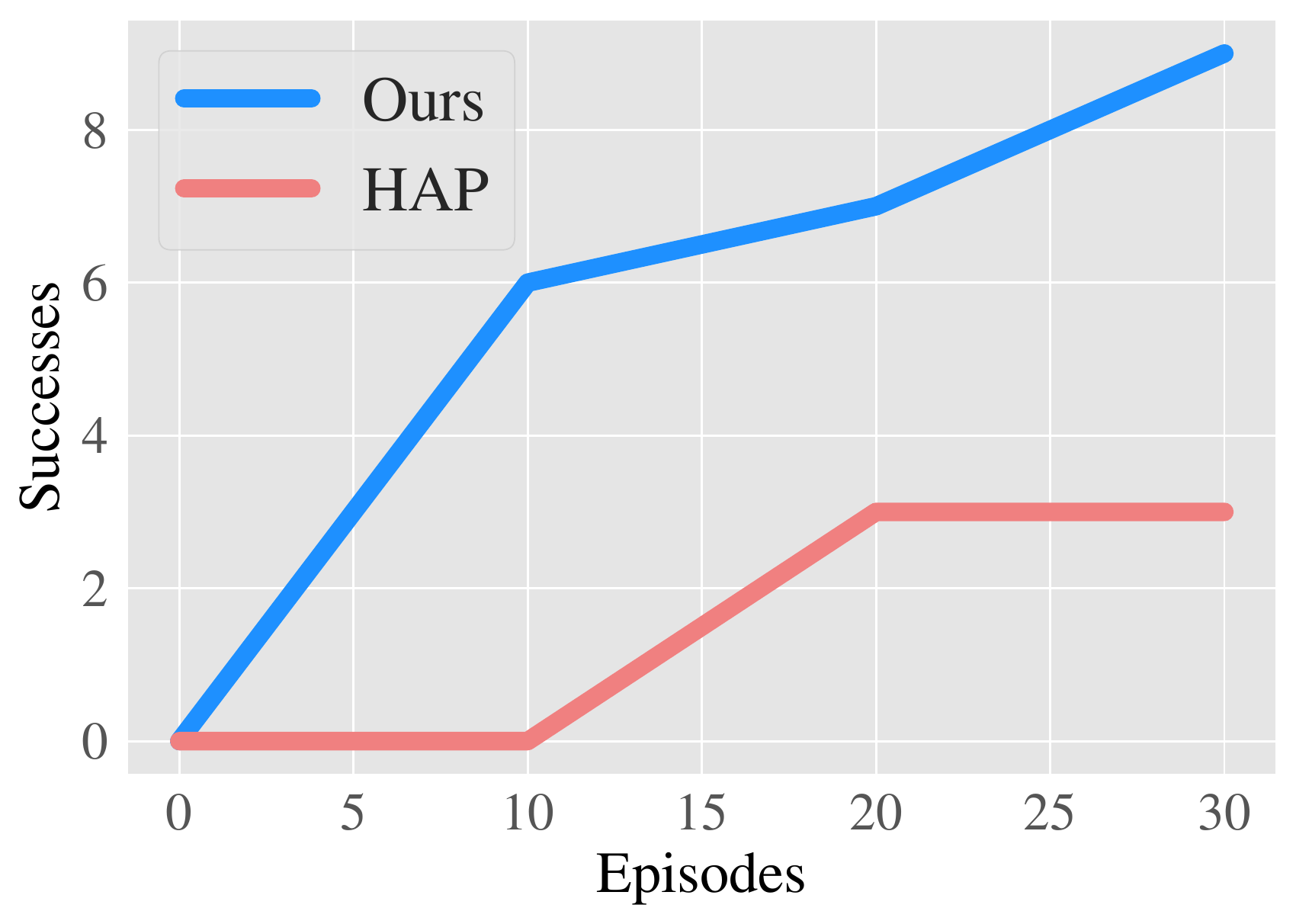}
    \vspace{-0.18in}
    \caption{Veggies}
    \end{subfigure}
    \vspace{-0.1in}
    \caption{\textbf{Action Space}: Success using DQN with the discretized action space, for reaching a specified goal image.}
    \label{fig:action-space}
    \vspace{-0.2in}
\end{figure}

We utilize visual affordances to create a discrete action space using a set of \cpoints and \dirs. We then train a Deep Q-Network (DQN) ~\cite{dqn} over this action space, for the above goal-conditioned learning problem.In \figref{fig:action-space}, we see that with \ours, the robot experiences more successes showing that a greater percentage of actions in the discretized action space correspond to meaningful object interactions.

\subsection{Analyzing Visual Representations}
Beyond showing better utility for robot learning paradigms, we analyze the quality of visual representations of the encoder learned in \ours. Two standard evaluations for this are (1) if they can help for downstream tasks and (2) how meaningful distances in their feature spaces are.

\begin{wraptable}{l}{.48\linewidth}
\vspace{-0.15in}
\hspace{-0.05in}
\resizebox{\linewidth}{!}{%
\begin{tabular}{lcc}
\toprule
 & \texttt{\ours} & \texttt{R3M} \cite{r3m}  \\ 
\midrule
\texttt{microwave} & \textbf{0.16} & 0.10 \\
\texttt{slide-door}  & \textbf{0.84}  & 0.70 \\
\texttt{door-open}  & \textbf{0.13} & 0.11 \\
\midrule
\bottomrule
\end{tabular}}
\vspace{-0.1in}
\caption{\footnotesize Behavior Cloning with \ours \vs R3M~\cite{r3m} representation. }
\vspace{-0.17in}
\label{tab:vis_rep_bc}
\end{wraptable}
\noindent\textbf{Finetuning}  To investigate if the visual representations are effective for control, we directly finetune a policy on top of the (frozen) visual encoder.  We evaluate on three simulated Franka environments, as shown in~\tabref{tab:vis_rep_bc}, and we see that \ours outperforms R3M on all tasks. (We finetuned the policy only for 2K steps, instead of 20K in the R3M paper).  This demonstrates that \ours visual representations contain information that is useful for control. 

\begin{figure}[t!]
    \centering
    \begin{subfigure}[b]{0.46\linewidth}
    \includegraphics[width=\linewidth]{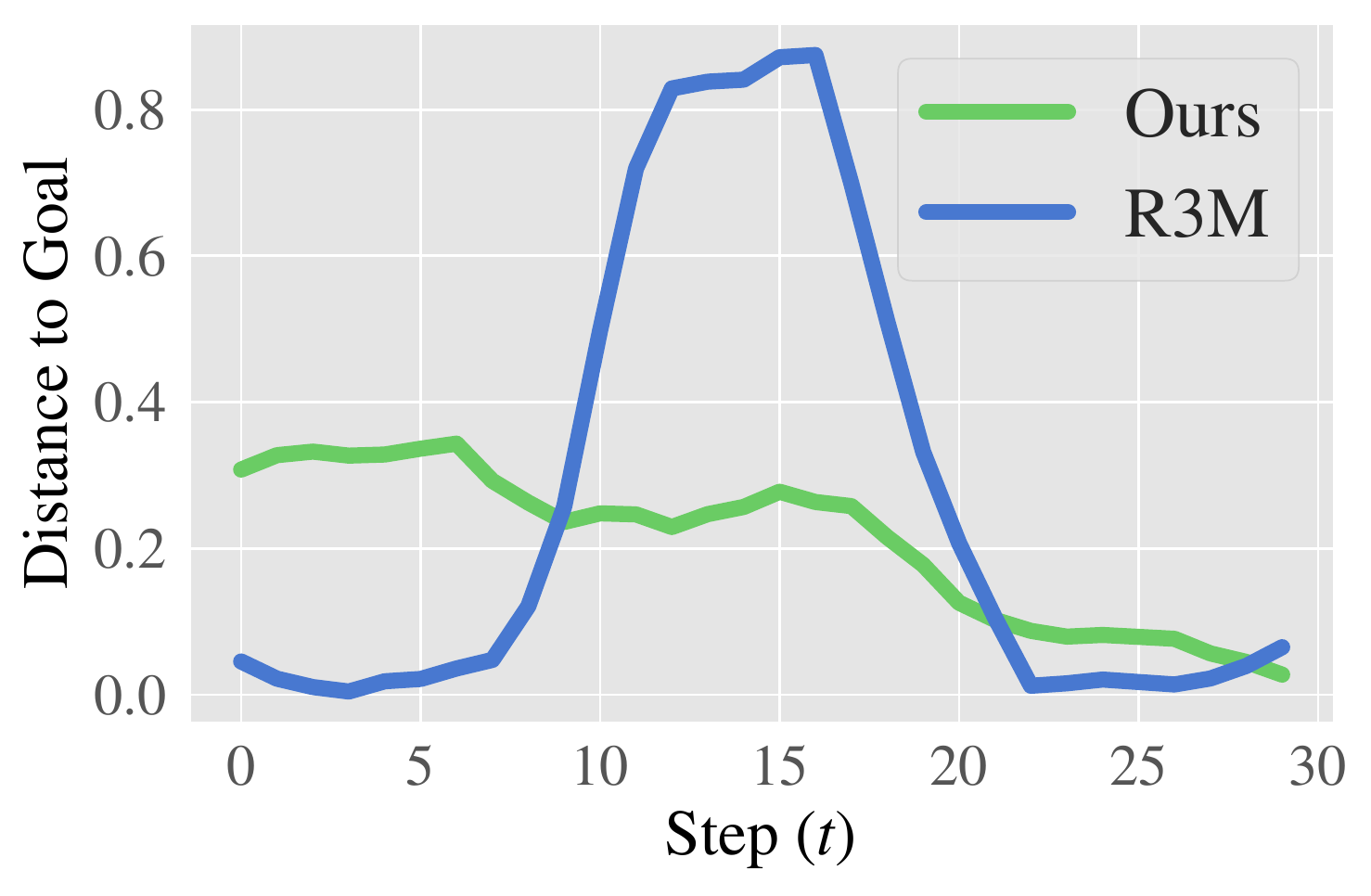}
    \end{subfigure}
    \quad
    \begin{subfigure}[b]{0.46\linewidth}
    \includegraphics[width=\linewidth]{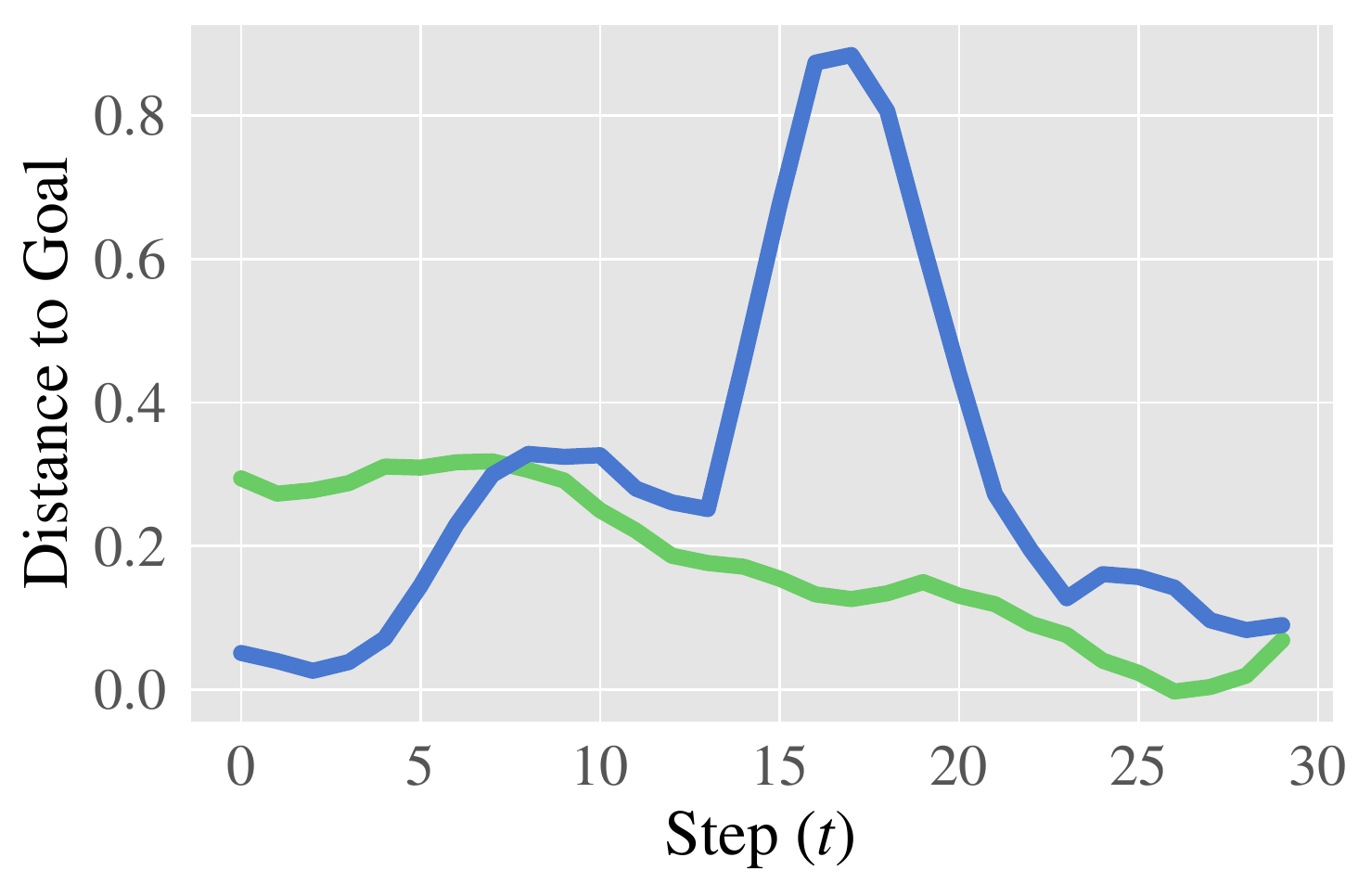}
    \end{subfigure}
    \vspace{-0.12in}
    \caption{
    \textbf{Feature space distance}: Distance to goal in feature space for \ours decreases monotonically for door opening. 
    }
    \vspace{-0.2in}
    \label{fig:feature_space_distance}
\end{figure}

\vspace{-0.15in}
\paragraph{Feature space distance} 
We record the distance in feature space between the current and goal image for every timestep in the episode, for both \ours and R3M~\cite{r3m} on successful cabinet opening trajectories. As shown in \figref{fig:feature_space_distance}, the distance for \ours decreases almost monotonically which correlates well with actual task progress.

\subsection{Failure Modes}
While \ours and the baselines see qualitatively similar successes, \ours in general sees a larger number of them and the \textit{average case} scenario for \ours is much better.  
 \begin{wrapfigure}{l}{.46\linewidth}
\vspace{-.1in}
\centering
\includegraphics[width=1\linewidth]{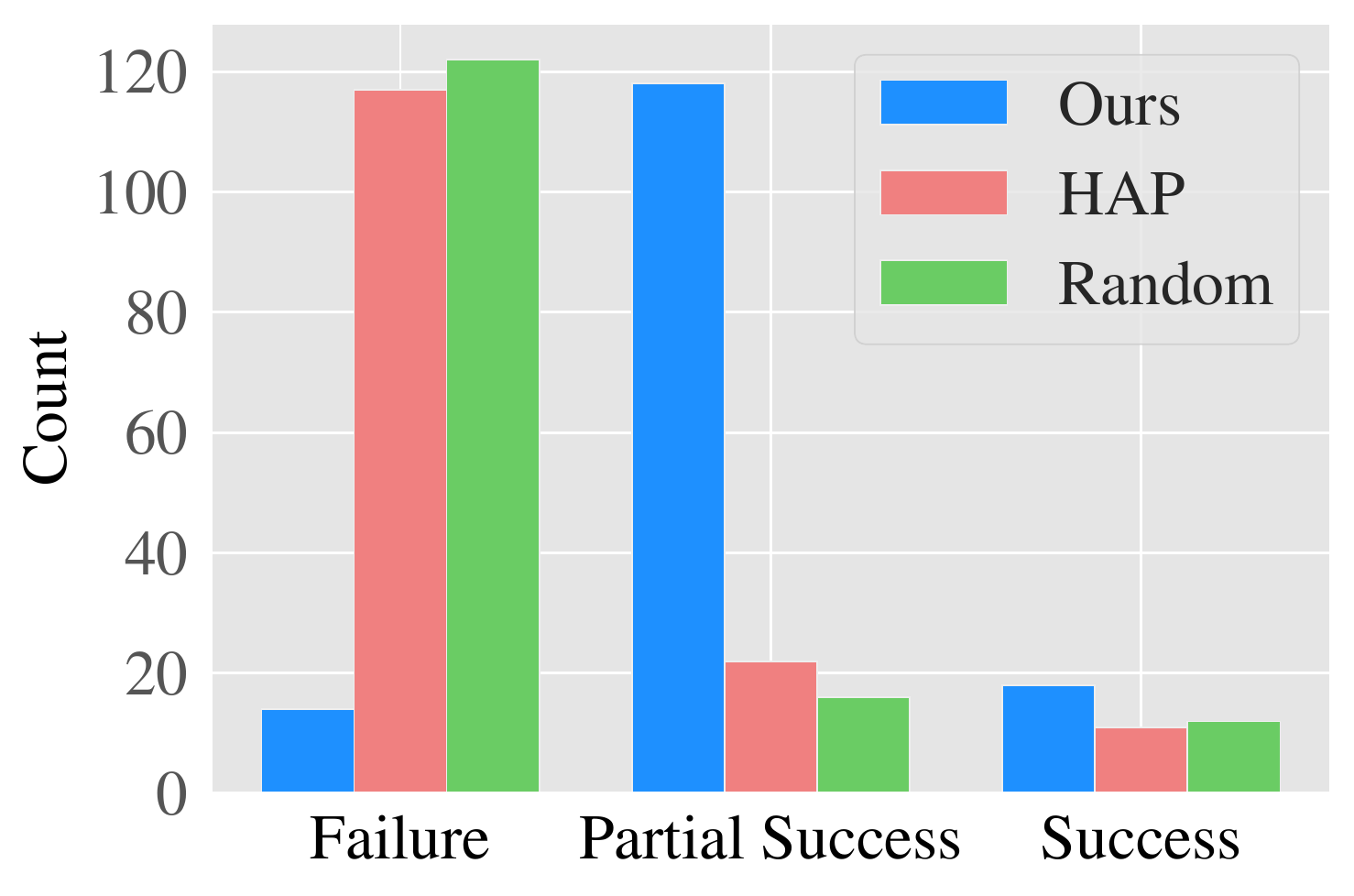}
\vspace{-.25in}
\caption{\footnotesize Failure mode analysis  }
\label{fig:failure-modes}
\vspace{-.12in}
\end{wrapfigure}
 For the cabinet opening task, we classify each collected episode into three categories: ``Failure", ``Partial Success" and ``Success". While \ours has a higher number of successful trajectories compared to the baselines (almost $2\times$), the number of partial successes is more than $6\times$ (\figref{fig:failure-modes}).

\section{Conclusion}
\label{sec:conc}
We propose Vision-Robotics Bridge (\ours), a scalable approach for learning useful affordances from passive human video data, and deploying them on many different robot learning paradigms (such as data collection for imitation, reward-free exploration, goal conditioned learning and paramterizing action spaces). Our affordance representation consists of contact points and post-contact trajectories. We demonstrate the effectiveness of this approach on the four paradigms and 10 different real world robotics tasks, including many that are in the wild. We run thorough experiments, spanning over 200 hours, and show that \ours drastically outperforms prior approaches. In the future, we hope to deploy on more complex multi-stage tasks, incorporate physical concepts such as force and tactile information, and investigate \ours in the context of visual representations.

\paragraph{Acknowledgements} We thank Shivam Duggal, Yufei Ye and Homanga Bharadhwaj for fruitful discussions and are grateful to Shagun Uppal, Ananye Agarwal, Murtaza Dalal and Jason Zhang for comments on early drafts of this paper. RM, LC, and DP are supported by NSF IIS-2024594, ONR MURI N00014-22-1-2773 and ONR N00014-22-1-2096.

{\small
\bibliographystyle{ieee_fullname}
\bibliography{main}
}

\clearpage
\appendix
\section*{Appendix}
\section{Result Videos}
\blfootnote{$^\star$equal contribution}
Several qualitative rollout videos are available at the \href{https://robo-affordances.github.io}{VRB website}.

\section{Affordance Model Setup}

\noindent{\textbf{Data Extraction: }} Our training setup involves learning from EpicKitchens-100 Videos \cite{EPICKITCHENS}. This dataset contains many hours of videos of humans performing different kitchen tasks. We use each sub-action video (such as `open door' or `put cup on table') as training sequences. Consider a video ($V$) consisting of $T$ frames, $V = \{I_1, ..., I_T\}$. Using 100 DOH annotations \cite{100doh} (available alongside the dataset), we find all of the hand-object contact points and frames for each hand in the video. As mentioned in Section 3, let  model output $f_\text{hand}(I_t) = \{h_t^l, h_t^r, o_t^l, o_t^r\}$, where $o^l$, $o^r$ are the contact variables and $h^l$, $h^r$ are the hand bounding boxes. We find the first contact timestep and select the active hand (left or right) as the hand side to consider for the whole trajectory. This is found by first binning $o_t$ and looking for all types that have contact with `Portable' or `Fixed' objects. These are assigned 1, while all others are assigned 0. We smooth the set of contact variables using a Savitzky–Golay filter \cite{savitzky1964smoothing} using a threshold of 0.75 (with window size 7). This should eliminate any spurious detections. We use the skin segmentation approach from \cite{hoi}, to find the contact points, $\{c_i\}^N$, at the contact timestep around the active hand. We then fit a GMM with $k = 5$ to the set of contact points to determine $\mu_1, ..., \mu_5$. We found that learning without a covariance, $\Sigma$, was more stable thus we only aim to learn the $\mu_1$. The input image becomes the first image before the contact where the hand is not visible. If the contact points or trajectory are not in the frame of this initial image (if the camera has moved), we then discard the trajectory. We use crops of size 150x150 (full image size is 456 x 256), which improves robustness at test time. We train on around 54K image-trajectory-contact point tuples. We include visualizations of the affordance model outputs on the \href{https://robo-affordances.github.io/}{VRB website}.

\noindent{\textbf{Architecture: }}
We use the ResNet18 encoder from ~\cite{r3m} as $g_\phi$, as our visual backbone. Our model has two heads, a trajectory head and a contact point head. We use the spatial features from the ResNet18 encoder (before the average pooling layer) as an input to three deconvolutional layers and two convolutional blocks with kernel sizes of 2 and 3 respectively, and channels: [256, 128, 64, 10, 5]. We use a spatial softmax to obtain $\hat{mu}_k$ for where $k = 1, ..., 5$. Our trajectory network is a transformer encoder with 6 self-attention layers with 8 heads each, and uses the output of the ResNet18 encoder (flattened), which has dimension 512. The output of the transformer encoder is used to predict a trajectory of length 5, using an MLP with two layers with hidden size 192.

\noindent{\textbf{Training: }} We train our model for 500 Epochs, using a learning rate of 0.0001 with cosine scheduling, and the ADAM \cite{kingma2014adam} optimizer. We train on 4 GPUs (2080Ti) for about 18 hours. 

\section{Robotics Setup}  

\noindent{\textbf{Hardware setup: }} For all the tasks we assume the following structure for robot control for each trajectory. We first sample a rotation configuration for the gripper. The arm then moves to the contact point $c$, closes its gripper, and moves to the points in the post-contact trajectory $\tau$. For the initial rotation of the Franka, joints 5 and 6 can take values in [0, 30, 45] degrees, while joint4 is fixed to be 0 degrees. For the Hello-Robot, the roll of the end-effector is varied in the range of [0, 45, 90] degrees. Once the orientation is chosen for the trajectory, we perform 3DOF end-effector control to move between points. Given two points a and b, we generate a sequence of waypoints between them to be reached using impedance control for the Franka. The Hello-Robot is axis aligned and has a telescoping arm, thus we did not need to build our own controller. We do not constrain the orientation to be exactly the same as what was selected in the beginning of the trajectory, since this might make reaching some points infeasible. For all tasks and methods we evaluate success rate by manual inspection of proximity to the goal image after robot execution (for imitation learning, goal reaching and affordance as an action space), and evaluate coincidental success for exploration using manual inspection of whether the objects noticeably move over the course of the robot's execution trajectory. We provide larger versions of the result plots of successes presented in the main paper in Figures~\ref{fig:goal-plots-supp} and~\ref{fig:expl-plots-supp}.

\noindent{\textbf{Affordance Model to Robot Actions:}} Reusing terminology from Section 3, the affordance model output is  $f_\theta(I_t) = \hat{p}_c, \hat{\tau}$, where $\hat{p}_c = \sum_{k=0}^K \alpha_k \mathcal{N}(\hat{\mu_k}, \hat{\Sigma_k})$, and $\hat{\tau} = \{w_i\}^M$. We can convert this into a 3D set of waypoints using a hand-eye calibrated camera, and obtain a 3D grasp point from $\hat{p}_c$, and a set of 3D waypoints from $\hat{\tau}$.

\noindent{\textbf{Imitation from Offline Data Collection:}} We use our affordance model to collect data for different tasks, and then evaluate whether this data can be used to reach goal images using $k$-NN and Behavior cloning. 
As mentioned in Sec 3.3.1, given an image $I_t$, the affordance model produces $(c, \tau) = f_\theta(I)$. In addition to storing $I_t$, $c$ and $\tau$, we also store the sequence of image observations (queried at a fixed frequency) seen by the robot when executing this trajectory $O_{1:k}$, where $k$ is the total number of images in the trajectory. $k$ varies across different trajectories (since it depends on $c$ and $\tau$). These intermediate images $O_i$ enable us to determine how close a trajectory is to the given goal image. For each trajectory, the distance to goal image $I_g$ is given by $\min_i || \psi(I_g) - \psi(O_i) ||_2^2 $, where $\psi$ is the R3M embedding space. We then use this distance to produce a set of K trajectories with smallest distances to the goal $I_g$. For $k$-NN, we simply run $(c, \tau)$ from each of these filtered trajectories. For Behavior cloning, we first train a policy that predicts $(c, \tau)$ given image $I$ using this set of trajectories, and then run the policy $\pi$ on the robot. We summarize this is Algorithm \ref{algo:im_from_dc}. We fix the number of top trajectories K to be 10 for $k$-NN and 20 for behavior cloning. The number of trajectories for initial data collection used for each task is listed in \ref{tab:num_traj}. For $k$-NN, the success is averaged across all K runs on the robot. For behavior-cloning, we parameterize the policy $\pi$ using a CVAE, where the image is the context, the encoder and decoder are 2 layer MLPs with 64 hidden units and the latent dimension is 4. During inference, we sample from the CVAE given the current image as context, and report success averaged across 10 runs. The quality of data collected by the robot using VRB which is used for imitation can be in seen in the videos on the
\href{https://robo-affordances.github.io/}{VRB website}.

Although many of our household object categories might be present in the videos of Epic-Kitchens \cite{EPICKITCHENS}, specific \textit{instances} of objects do not appear in training, thus every object our approach is evaluated on is new. To test generalization to ``rare'' (held-out) objects and evaluate the grasping success using \ours's affordances, see Table~\ref{tab:newobjects}. \ours consistently outperforms our most competitive baseline, Hotspots \cite{hotspots}. 

\begin{table}[t]
    \centering
    \resizebox{0.7\linewidth}{!}{%
    \begin{tabular}{c|c|c}
        \toprule
        \textbf{Object} & \textbf{VRB} & \textbf{Hotspots}\\
        \midrule
        VR Controller &  \textbf{0.27} & 0.13 \\
        Chain & \textbf{0.33} & 0.20 \\
        Hat & 0.07 & \textbf{0.20} \\
        Tape & \textbf{0.13} & 0.00 \\
        Cube & 0.00 & 0.00 \\
        Sanitizer & \textbf{0.27} & 0.20 \\
        Stapler & \textbf{0.53} & 0.20 \\
        Shoe & \textbf{0.33} & 0.13 \\
        Mouse & \textbf{0.27} & 0.00 \\ 
        Hair-Clip & \textbf{0.47} & 0.20 \\
        \bottomrule
    \end{tabular}
    }
    \caption{\ours for grasping held-out ``rare'' objects}
    \label{tab:newobjects}
\end{table}

\begin{algorithm}[t!]
\caption{Imitation from Offline Data Collection}
\label{alg:method}
\begin{algorithmic}[1]

\REQUIRE Dataset of trajectories \{($I_t$, $O_{1:k}$, $c$, $\tau$)\}
\REQUIRE Number of top trajectories K
\REQUIRE Goal Image $I_g$
\REQUIRE R3M embedding space $\psi$

\STATE For each trajectory $\mathcal{T}$, compute \\ $d_\mathcal{T} =\min_i || \psi(I_g) - \psi(O_i) ||_2^2 $
\STATE Rank trajectories in ascending order of $d_\mathcal{T}$. Create set $\mathcal{K}$ = \{$(c, \tau)$\} of the top K ranked trajectories. 
\IF{\textbf{$k$-NN}}
    \STATE Execute $\mathcal{K}$ on the robot.
\ELSE
    \STATE Assert \textbf{behavior cloning}
    \STATE Train a policy $\pi(c, \tau|I)$ using  $\mathcal{K}$. \STATE Execute $c, \tau \sim \pi(.|I)$ on the robot. 
\ENDIF
\\[1ex]
\end{algorithmic}
\vspace{-0.05in}
\label{algo:im_from_dc}
\end{algorithm}

\begin{algorithm}[t!]
\caption{Exploration / Goal Reaching}
\label{alg:method}
\begin{algorithmic}[1]
\REQUIRE Number of iterations J
\REQUIRE Number of top trajectories K
\REQUIRE Number of initial trajectories $N_0$, \\and for subsequent fitting iterations $N_s$ 
\REQUIRE Affordance model $f_\theta$
\REQUIRE Tradeoff probability $p$
\REQUIRE Visual change model $\Phi$ (only for \textbf{exploration})
\REQUIRE R3M embedding $\psi$ (only for \textbf{goal reaching})
\REQUIRE Goal Image $I_g$ (only for \textbf{goal reaching})
\STATE \textbf{initialize:} World model $\mathcal{M}$,
Replay buffer $\mathcal{D}$,
\STATE Execute $(c, \tau) = f_\theta(I)$ on the robot for $N_0$ iterations to collect initial dataset $\mathcal{D}$ = \{($I$, $O_{1:k}$, $c$, $\tau$)\}
\FOR{iteration 1:J}
\STATE For each trajectory $\mathcal{T}_{0:k}$, compute \\
\IF{\textbf{exploring}}
    \STATE compute $\text{EC}_\mathcal{T} = ||\phi(O_{1}) - \phi(O_{k})||_2$
    \STATE Rank trajectories in descending order of $\text{EC}_\mathcal{T}$
\ELSE
    \STATE Assert \textbf{goal reaching}
    \STATE compute $\text{d}_\mathcal{T} = \min_i ||\psi(I_{g}) - \psi(O_{i})||_2$
    \STATE Rank trajectories in ascending order of $d_\mathcal{T}$
\ENDIF
\STATE Create set $\mathcal{K}$ = \{$(c, \tau)$\} of top K ranked trajectories. 
\STATE Compute $\hat{c}, \hat{\tau}$ = mean($\mathcal{K}$)
\STATE For $N_s$ iterations, set $(c, \tau)$ = $f_\theta(I)$ with probability $p$, otherwise  set $(c, \tau) = (\hat{c}, \hat{\tau})$.
\STATE Execute $(c, \tau)$ on the robot and append data to $\mathcal{D}$
\ENDFOR
\\[.7ex]
\end{algorithmic}
\vspace{-0.05in}
\label{algo:expl_goal_reaching}
\end{algorithm}

\begin{algorithm}[t!]
\caption{Affordance as Action Space}
\label{alg:method}
\begin{algorithmic}[1]

\REQUIRE Affordance Model $f_\theta$
\REQUIRE Number of initial queries $q$
\REQUIRE Number of clusters for $c$, $N_c$ and for $\tau$, $N_\tau$
\REQUIRE Goal Image $I_g$
\REQUIRE RL algorithm with discrete action-space $RLA$
\REQUIRE R3M embedding space $\psi$

\STATE Query $f_\theta$ on the image of the scene q times \\to obtain a dataset \{$(c, \tau$\}
\STATE Fit a GMM $G_c$ with $N_c$ centers to $\{c\}$, and \\ a GMM $G_\tau$ and $N_\tau$ centers to $\{\tau\}$  

\STATE Create mapping $\mathcal{M}$ from $\mathcal{A}$ = [1..$N_c$*$N_\tau$] to values in the cross-product space of the centers of $G_c$ and $G_\tau$
\STATE Initialize Dataset $\mathcal{D}$ = \{\}, and $RLA$ with discrete action space $\mathcal{A}$ and random policy $\pi$.
\STATE Run \textbf{Sampling} and \textbf{Training} asynchronously
\WHILE{\textbf{Sampling}}
    \STATE Run $\pi$ on the image to get $a_d$. 
    \STATE $(c, \tau) = \mathcal{M}(a_d)$, execute on the robot and collect initial and final images $I_0$ and $I_T$
    \STATE Compute reward $r = ||\psi(I_T) - \psi(I_g)||_2$.
    \STATE Store  $(\psi(I_0), a_d,  \psi(I_T), r )$ in $\mathcal{D}$
\ENDWHILE
\WHILE{\textbf{Training}}
    \STATE  Sample data $\sim \mathcal{D}$, pass to $RLA$ for\\ training and updating $\pi$.
\ENDWHILE
\\[1ex]
\end{algorithmic}
\vspace{-0.05in}
\label{algo:action_space}
\end{algorithm}

\begin{table}[t!]
\centering
\resizebox{1\linewidth}{!}{%
\Huge
\begin{tabular}{lcccccccc}
\toprule
 & \textbf{Cabinet} & \textbf{Knife} & \textbf{Veg} & \textbf{Shelf}  & \textbf{Pot}  & \textbf{Door} & \textbf{Lid} & \textbf{Drawer} \\
\midrule
\texttt{$N_0$}  & 150 & 100 & 50 & 50 & 50 & 50 & 30 & 40 \\
\texttt{$N_s$}  & 50  & 50  & 30 & 30 & 30 & 50 & 30 & 40  \\
\bottomrule
\end{tabular}}
\caption{ Number of trajectories collected for various tasks, for Initial Data Collection ($N_0$) and for each subsequent fitting iteration for either goal reaching or exploration ($N_s$) }
\label{tab:num_traj}
\vspace{-0.2in}
\end{table}

\begin{figure*}[t!]
    \centering
    \begin{subfigure}[b]{0.23\linewidth}
    \includegraphics[width=\linewidth]{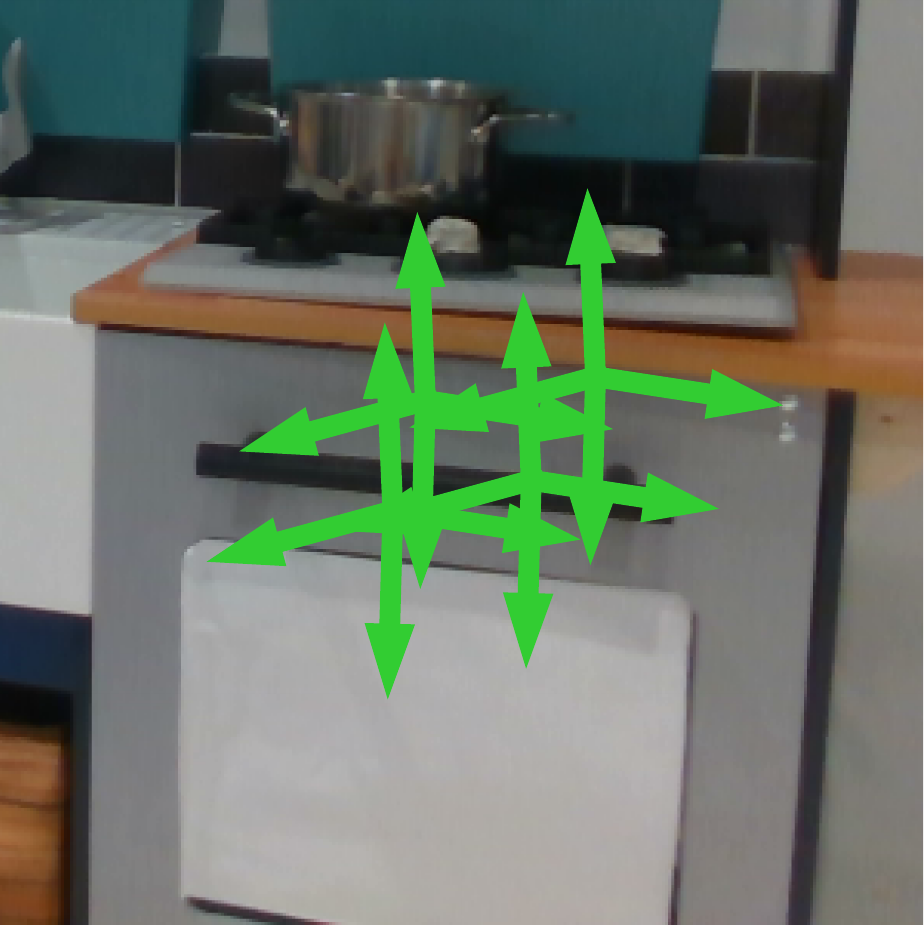}
    \caption{VRB - Cabinet}
    \end{subfigure}
    \begin{subfigure}[b]{0.23\linewidth}
    \includegraphics[width=\linewidth]{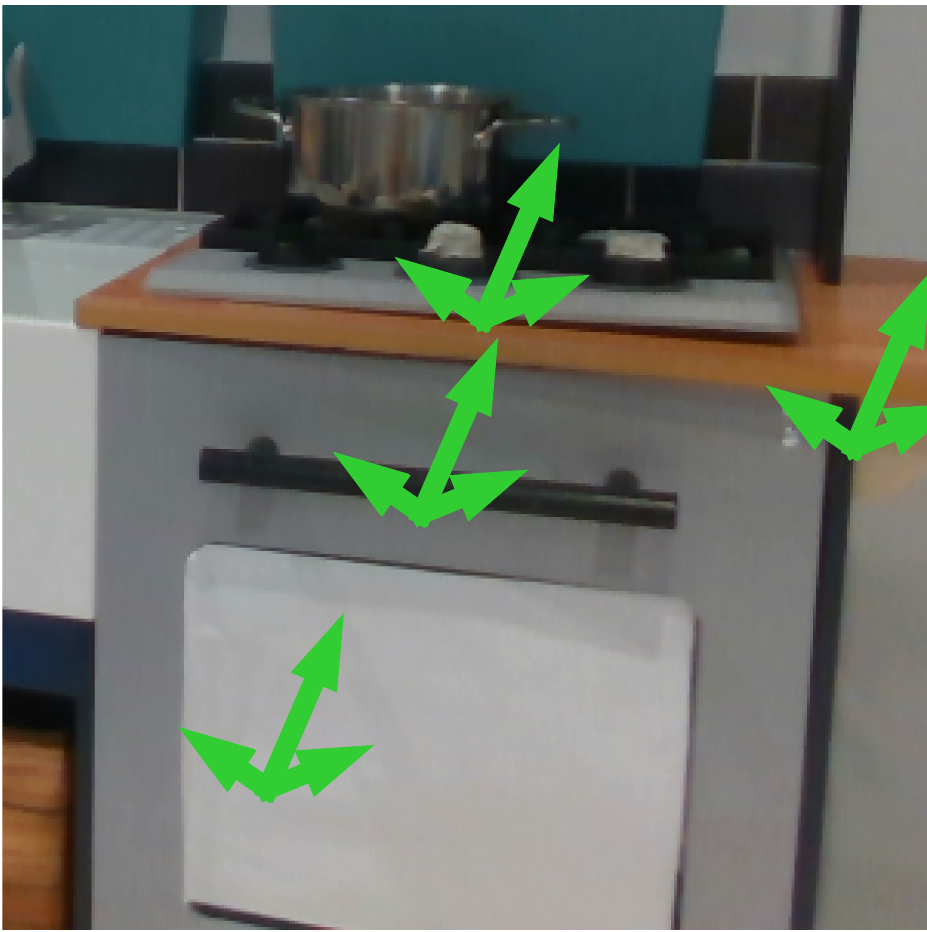}
    \caption{HAP - Cabinet}
    \end{subfigure}
     \begin{subfigure}[b]{0.26\linewidth}
    \includegraphics[width=\linewidth]{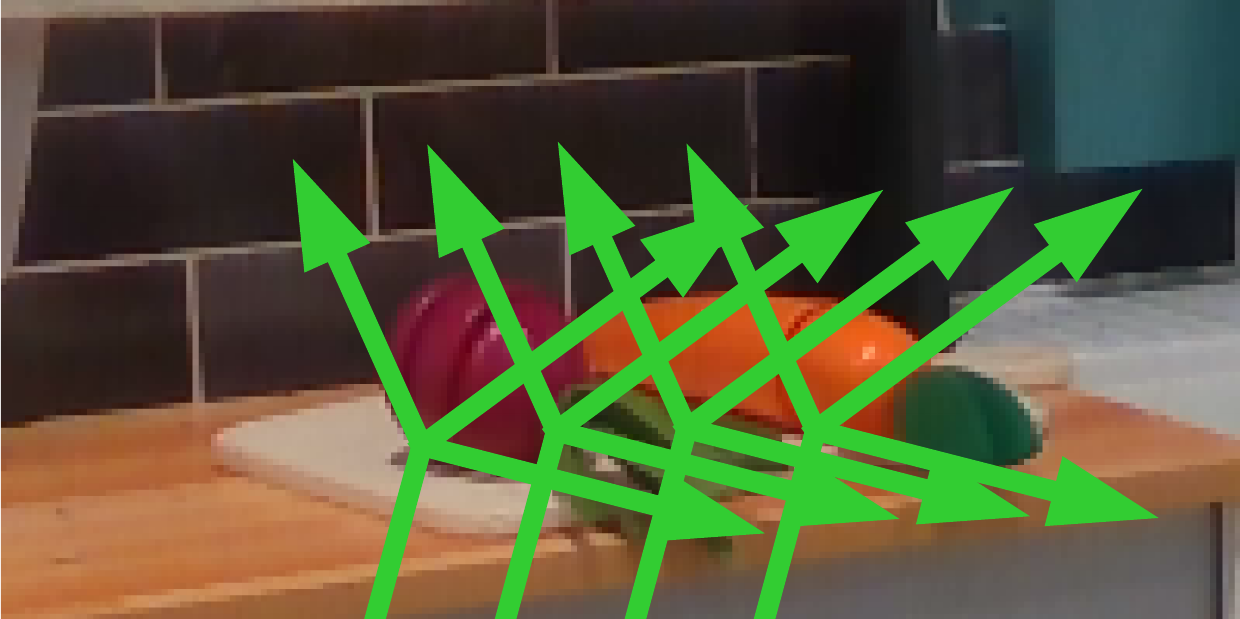}
    \caption{VRB - Veggies}
    \end{subfigure}
     \begin{subfigure}[b]{0.26\linewidth}
    \includegraphics[width=\linewidth]{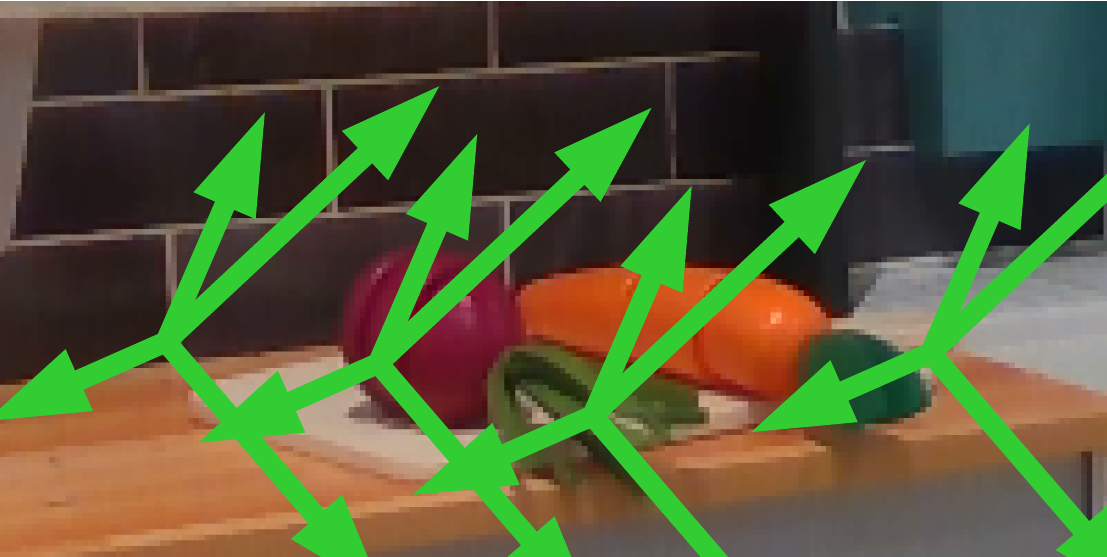}
    \caption{HAP - Veggies}
    \end{subfigure}
    \caption{Visualization for Affordance as an Action Space for VRB and HAP \cite{hap}, on the Cabinet and Veggies Tasks}
    \label{fig:aff_act_space_vis}
\end{figure*}

\noindent{\textbf{Exploration \& Goal Reaching: }} We apply our affordance model in the paradigms of exploration as well as goal reaching, where the robot uses the collected data to improve its behavior. As described in Section 3.3, we use a \emph{environment change} visual model to obtain intrinsic reward for exploration, while for goal-reaching we use \emph{distance to the goal} in a feature space like the R3M embedding space. For exploration, we want to \emph{maximize} the change between the first and last images of the trajectory, since greater perturbation of objects can lead to the discovery of useful manipulation skills. For goal-reaching, we \emph{minimize} the distance between the trajectory and the goal image, since this achieves the desired object state. In each case (exploration and goal-reaching), we rank the trajectories in the dataset using the appropriate metric, and then fit $(\hat{c}, \hat{\tau})$ to the \{$(c, \tau)$\} values of the top ranked trajectories. For subsequent data collection iterations, we use the affordance model $f_\theta$ with some probability $p$, but otherwise use $(\hat{c}, \hat{\tau})$ for execution on the robot. The newly collected data is then aggregated with the dataset, and the entire process repeated. We present this procedure in Algorithm \ref{algo:expl_goal_reaching}. The number of initial trajectories $N_0$ and trajectories for subsequent iterations $N_s$ for different tasks are listed in \ref{tab:num_traj}. For all experiments, we set $p$ = 0.35,  K = 10, J = 2. We include videos on the \href{https://robo-affordances.github.io/}{VRB website}, which show that as our system sees more data, its performance improves for both exploration and goal-reaching.

\textbf{Intrinsic Reward Model} We train a visual model which given a pair of images $(I_i, I_j)$, produces a binary image that captures how \emph{objects} move, and is not affected by changes in the robot arm or body position. Specifically, this model comprises the following -

\begin{equation}
\begin{aligned}
    \phi(I_i, I_j ) = g(||m(I_i) - m(I_j)||_2 ,  \\
    || \Psi(m(I_i)) - \Psi( m(I_j))||_2)
\end{aligned}
\label{eq:ec_rew}
\end{equation}

Here $m$ is a masking network which removes the robot from the image. We train this using around 100-200 hand-annotations of the robot in various scenes, and use this data to finetune a pretrained segmentation model $\Psi$ \cite{he2017mask}. We evaluate the l2-losses above only on \textbf{non-masked} pixels. Further, we also take into account distance in the feature space of the segmentation model to reduce sensitivity to spurious visual artifacts. The function $g$ applies heurestics including gaussian blurring to reduce effects of shadows, and a threshold for the change at each pixel, to limit false positives.

\begin{figure*}[h!]
    \centering
    \begin{subfigure}[b]{0.32\linewidth}
    \includegraphics[width=\linewidth]{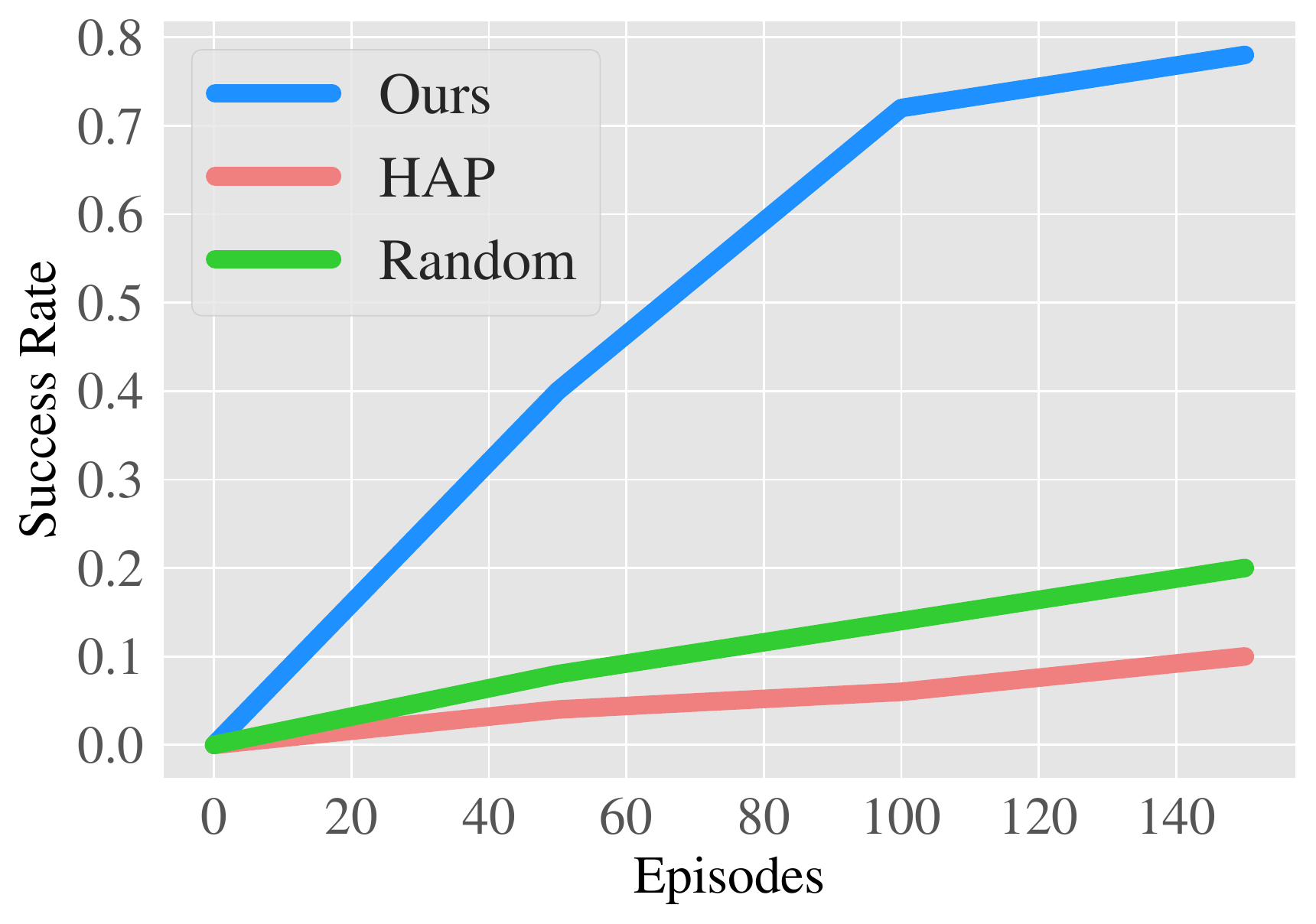}
    \caption{Door}
    \end{subfigure}
    \begin{subfigure}[b]{0.32\linewidth}
    \includegraphics[width=\linewidth]{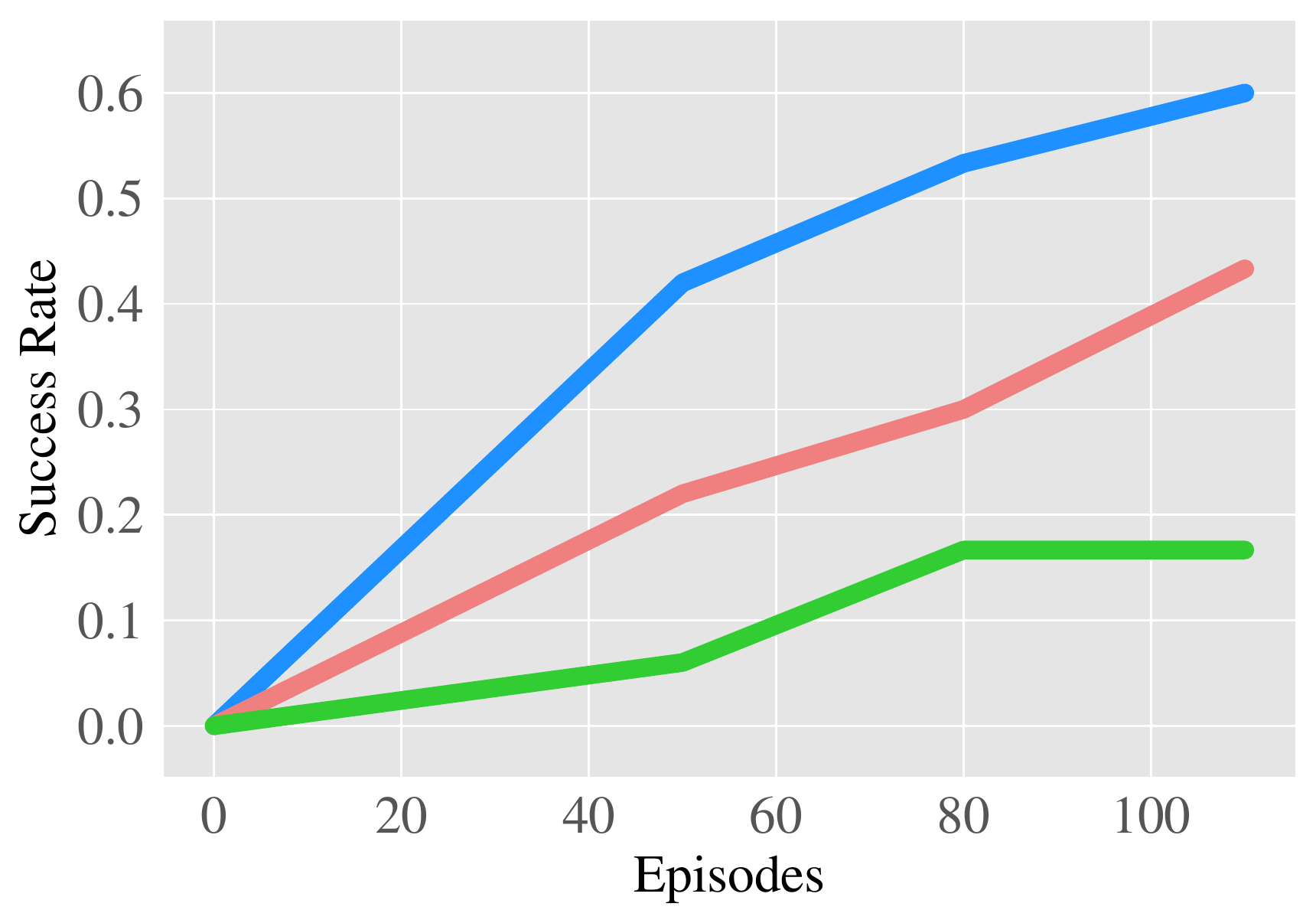}
    \caption{Veggies}
    \end{subfigure}
    \begin{subfigure}[b]{0.32\linewidth}
    \includegraphics[width=\linewidth]{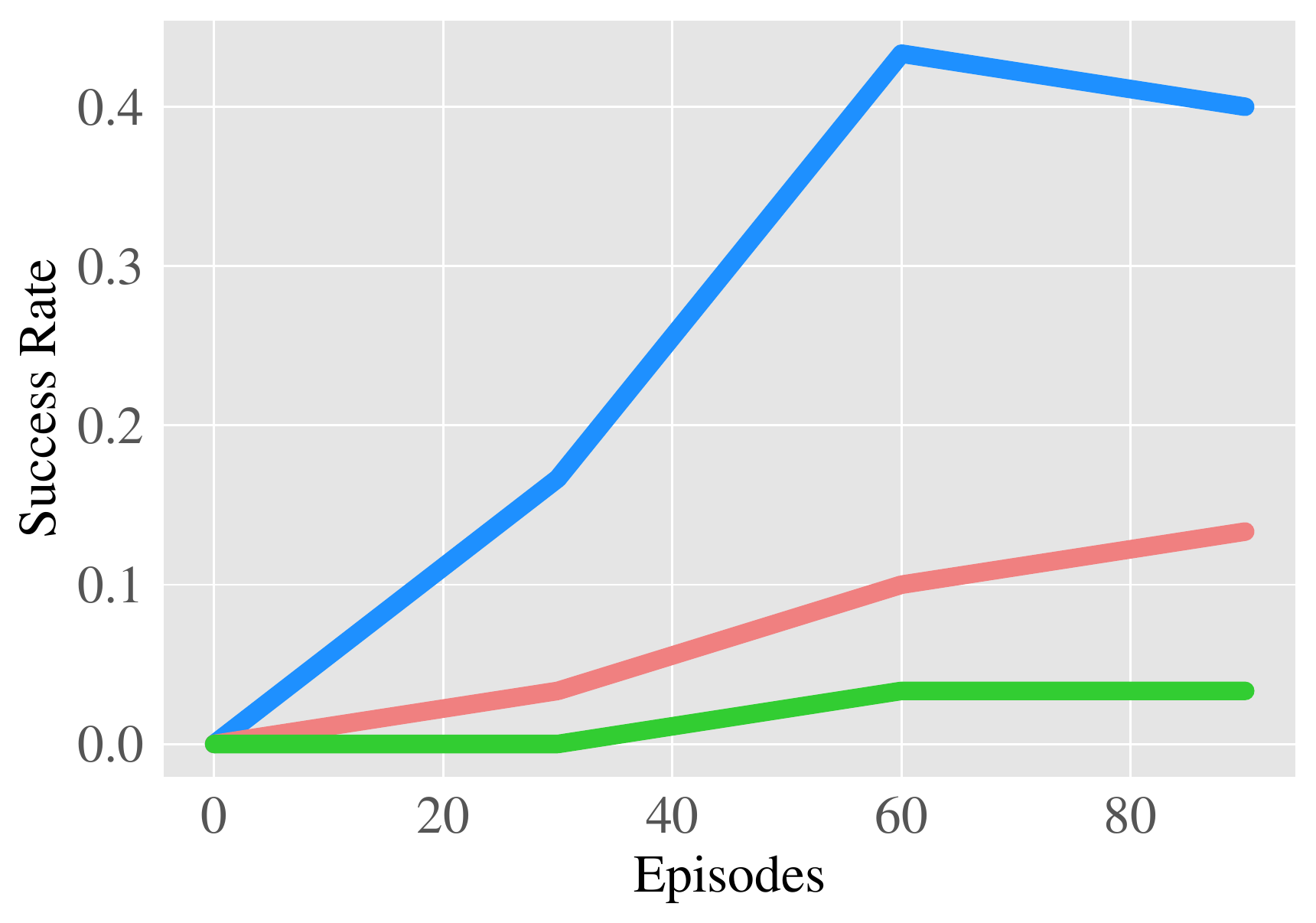}
    \caption{Lid}
    \end{subfigure}
    \begin{subfigure}[b]{0.32\linewidth}
    \includegraphics[width=\linewidth]{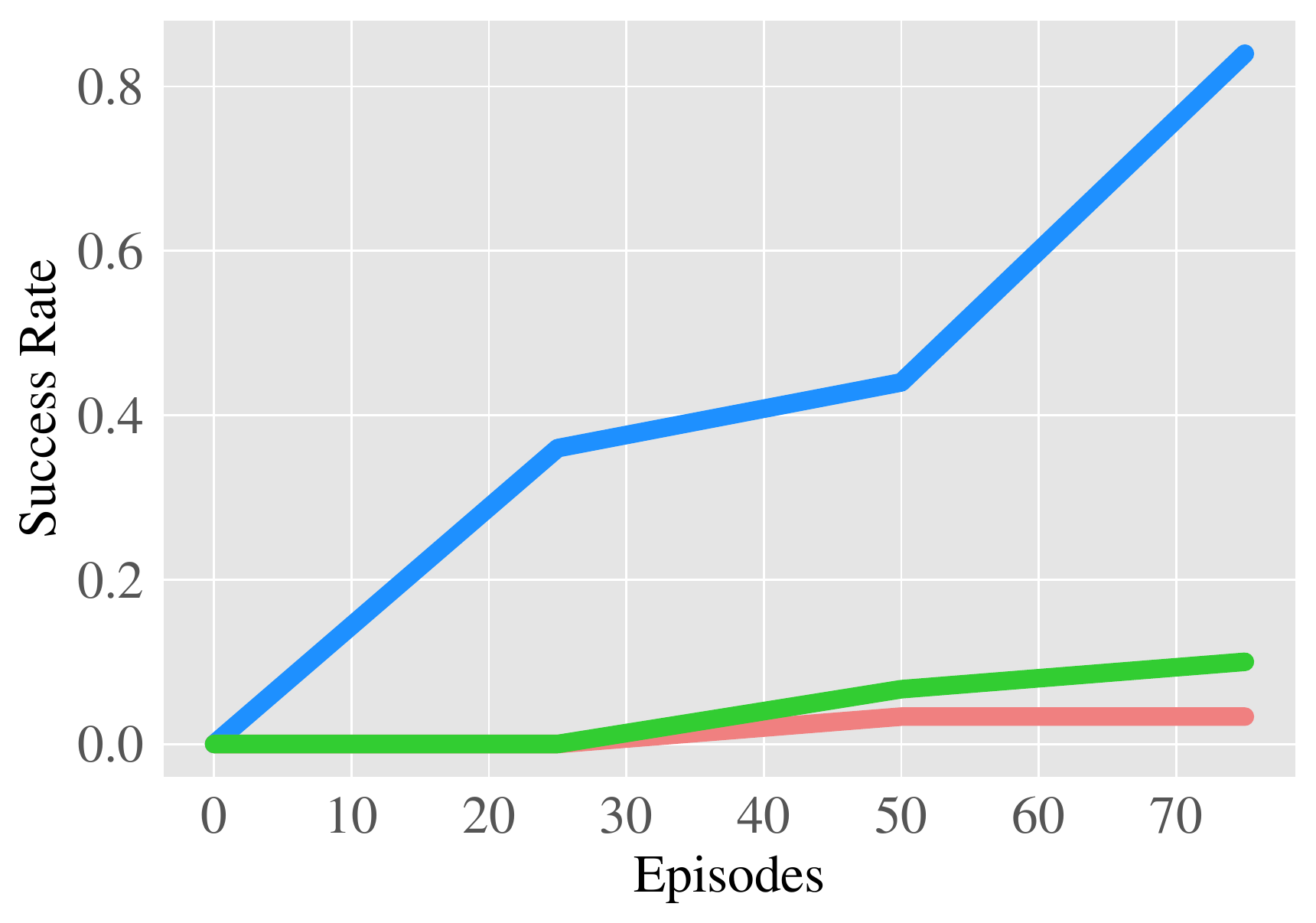}
    \caption{Dishwasher}
    \end{subfigure}
    \begin{subfigure}[b]{0.32\linewidth}
    \includegraphics[width=\linewidth]{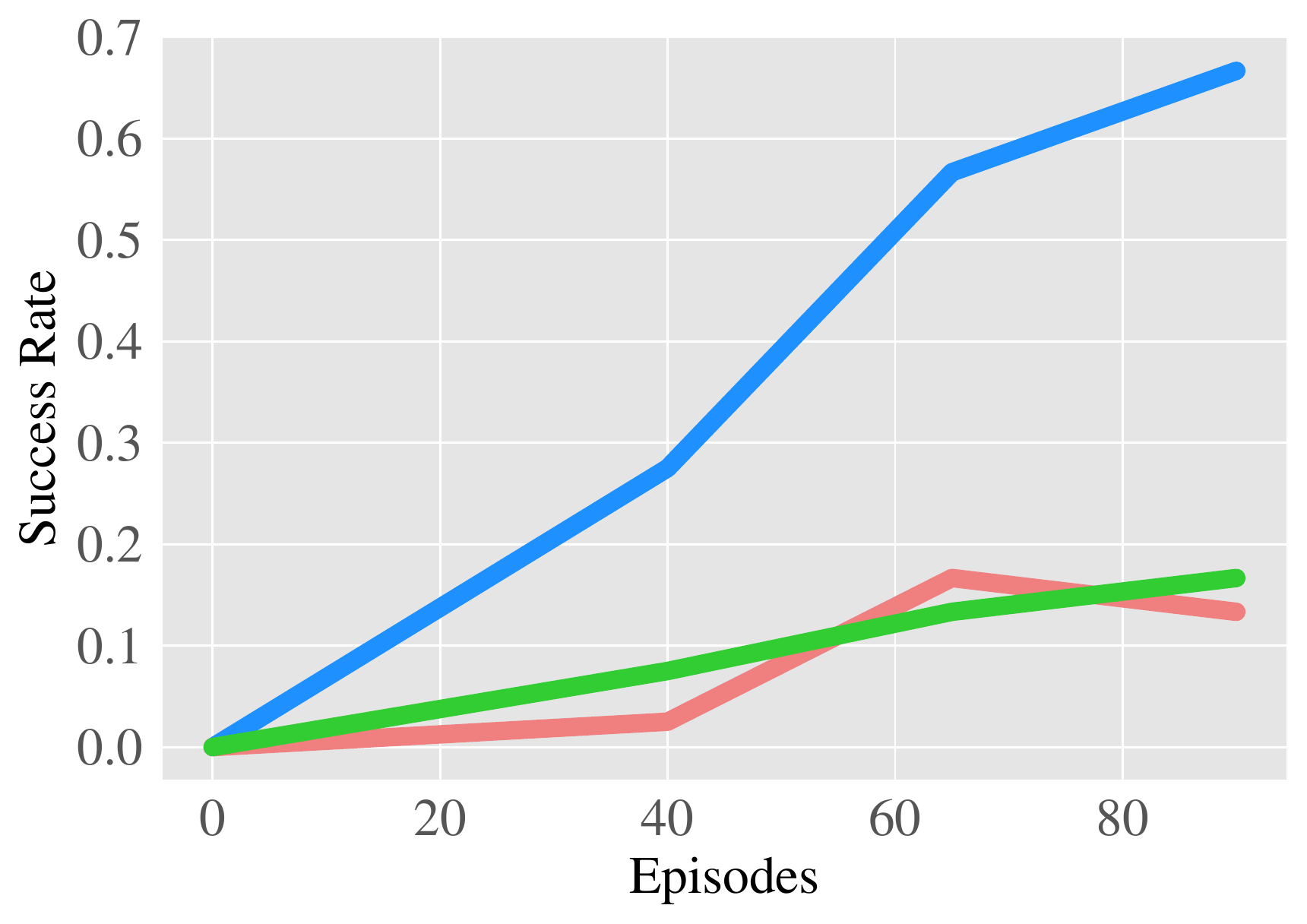}
    \caption{Drawer}
    \end{subfigure}
    \begin{subfigure}[b]{0.32\linewidth}
    \includegraphics[width=\linewidth]{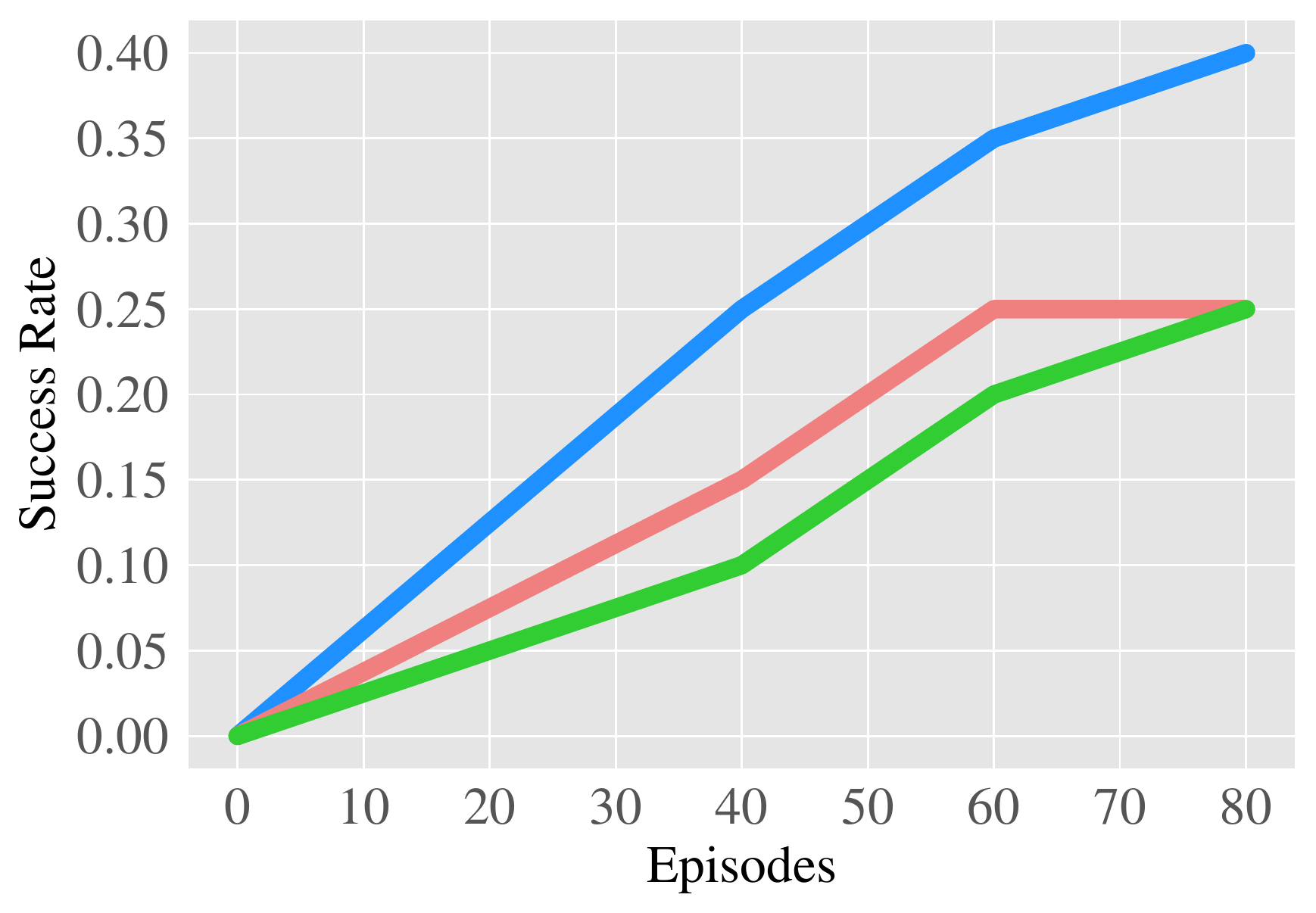}
    \caption{Garbage Can}
    \end{subfigure}
    \vspace{-0.1in}
    \caption{\textbf{Goal-conditioned Learning}:
    Success rate for reaching goal configuration for six different tasks. Sampling via \ours leads to faster learning and better final performance.
}    
    \vspace{-0.1in}
    \label{fig:goal-plots-supp}
\end{figure*}

\begin{figure*}[t!]
    \centering
    \begin{subfigure}[b]{0.32\linewidth}
    \includegraphics[width=\linewidth]{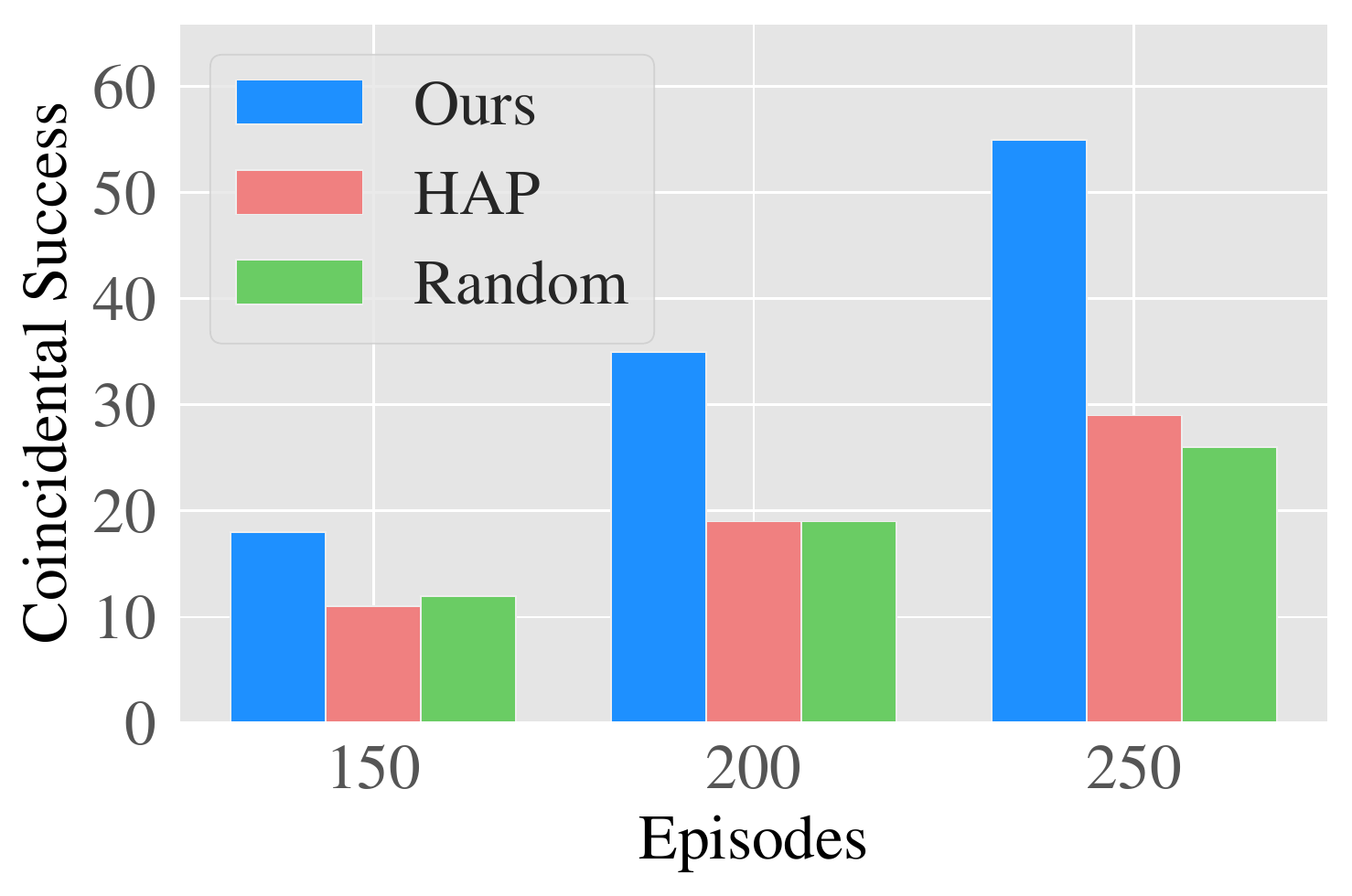}
    \caption{Cabinet}
    \end{subfigure}
    \begin{subfigure}[b]{0.32\linewidth}
    \includegraphics[width=\linewidth]{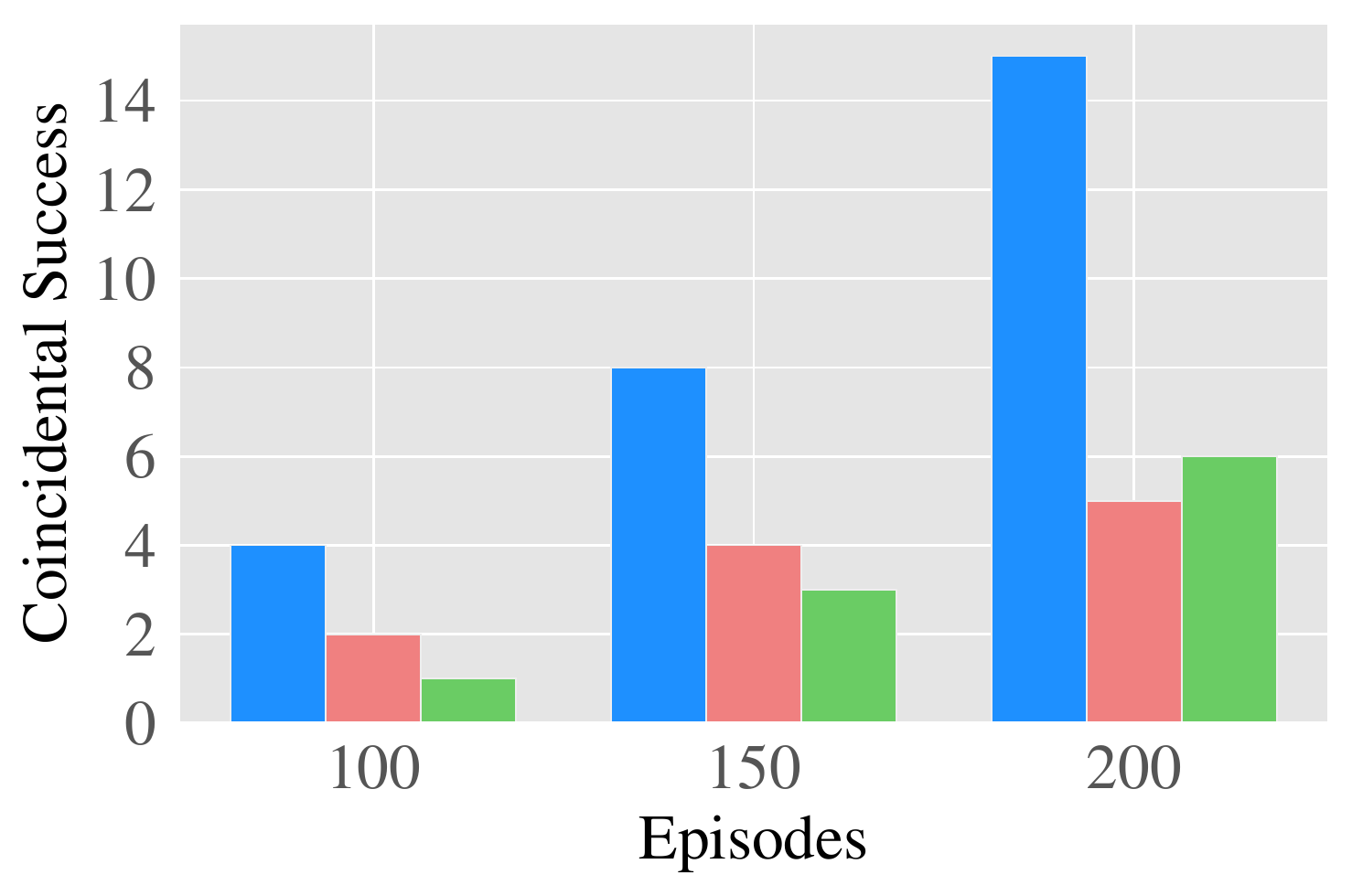}
    \caption{Knife}
    \end{subfigure}
    \begin{subfigure}[b]{0.32\linewidth}
    \includegraphics[width=\linewidth]{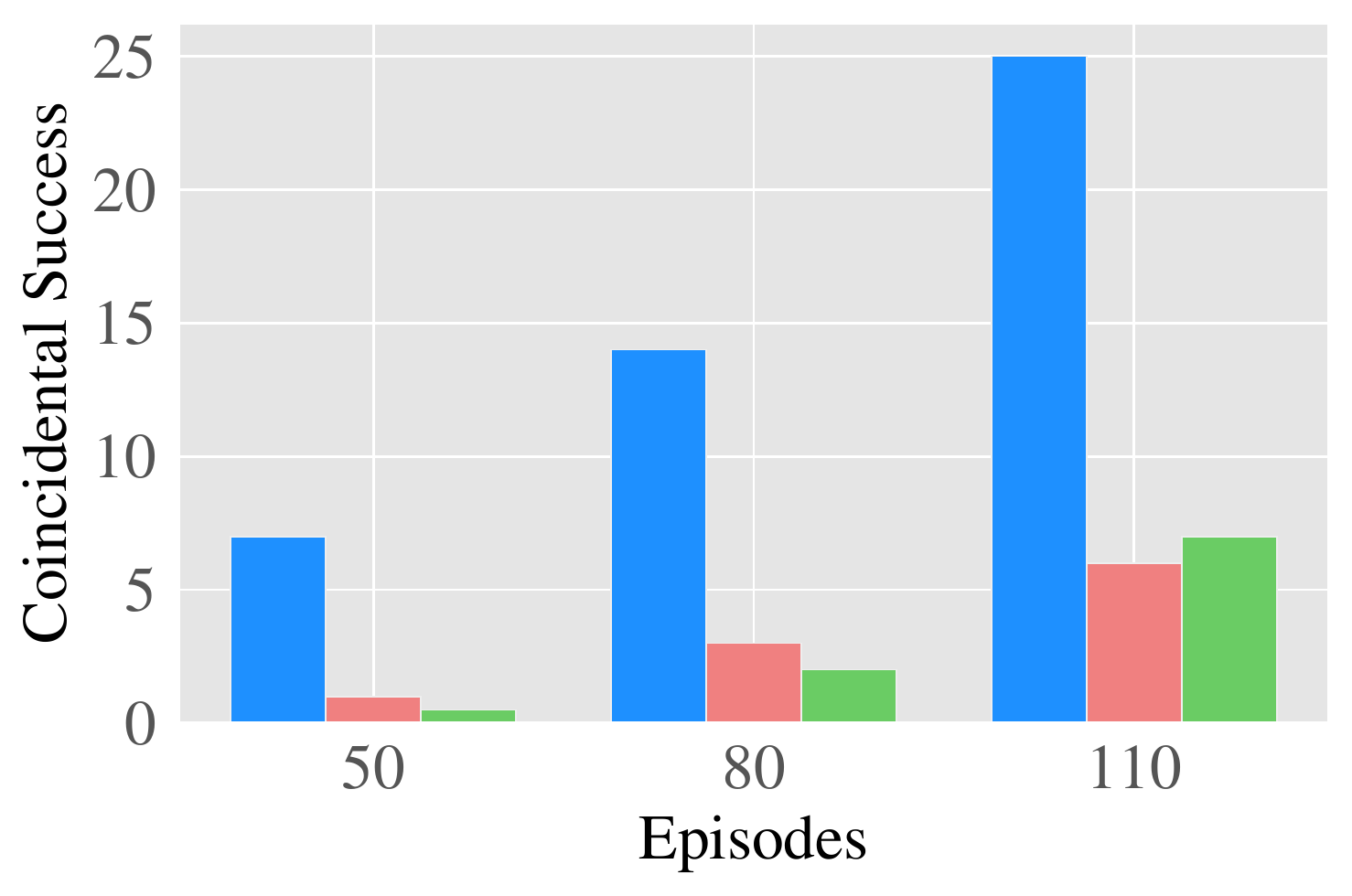}
    \caption{Stovepot}
    \end{subfigure}
    \begin{subfigure}[b]{0.32\linewidth}
    \includegraphics[width=\linewidth]{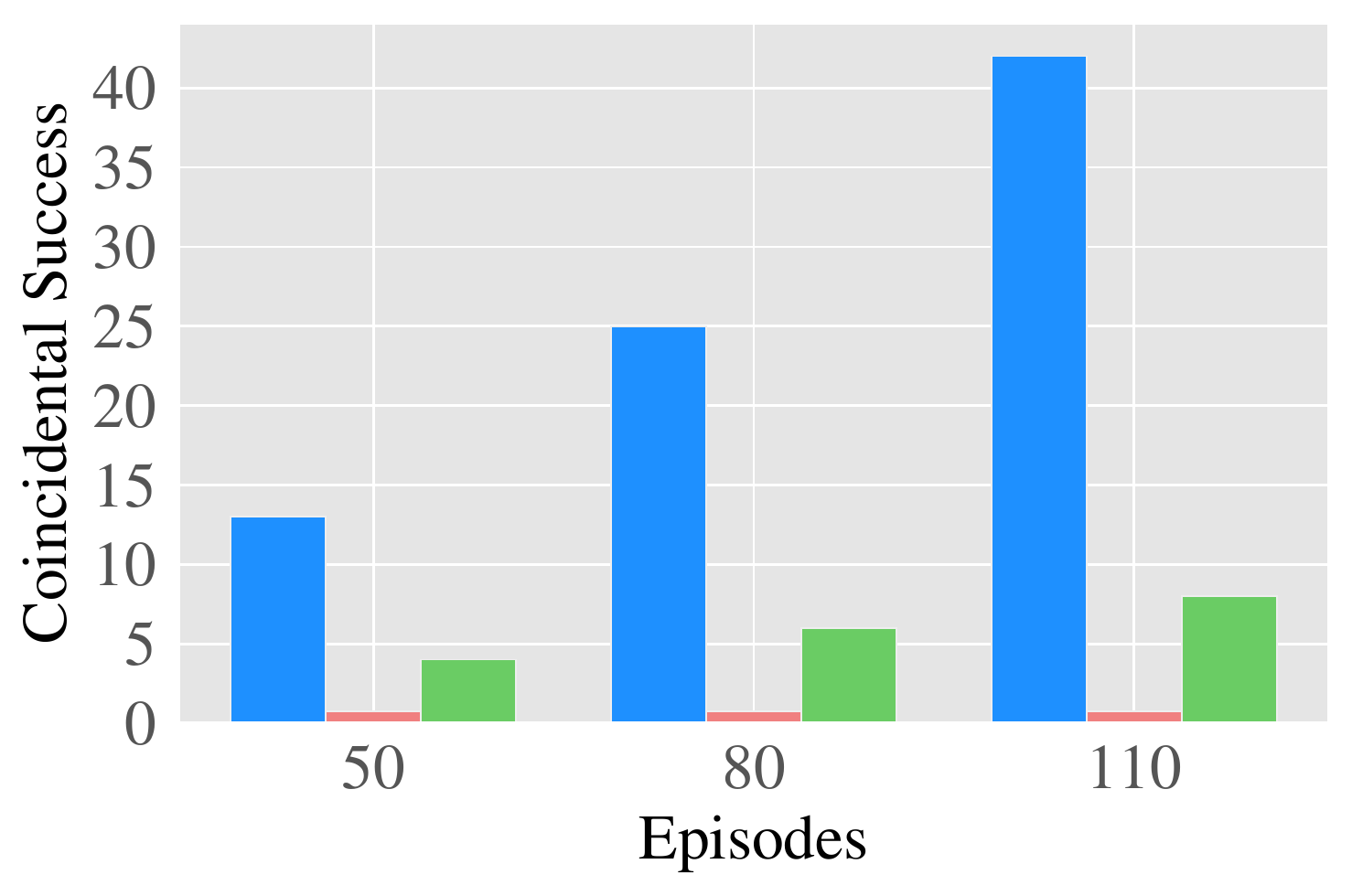}
    \caption{Shelf}
    \end{subfigure}
    \begin{subfigure}[b]{0.32\linewidth}
    \includegraphics[width=\linewidth]{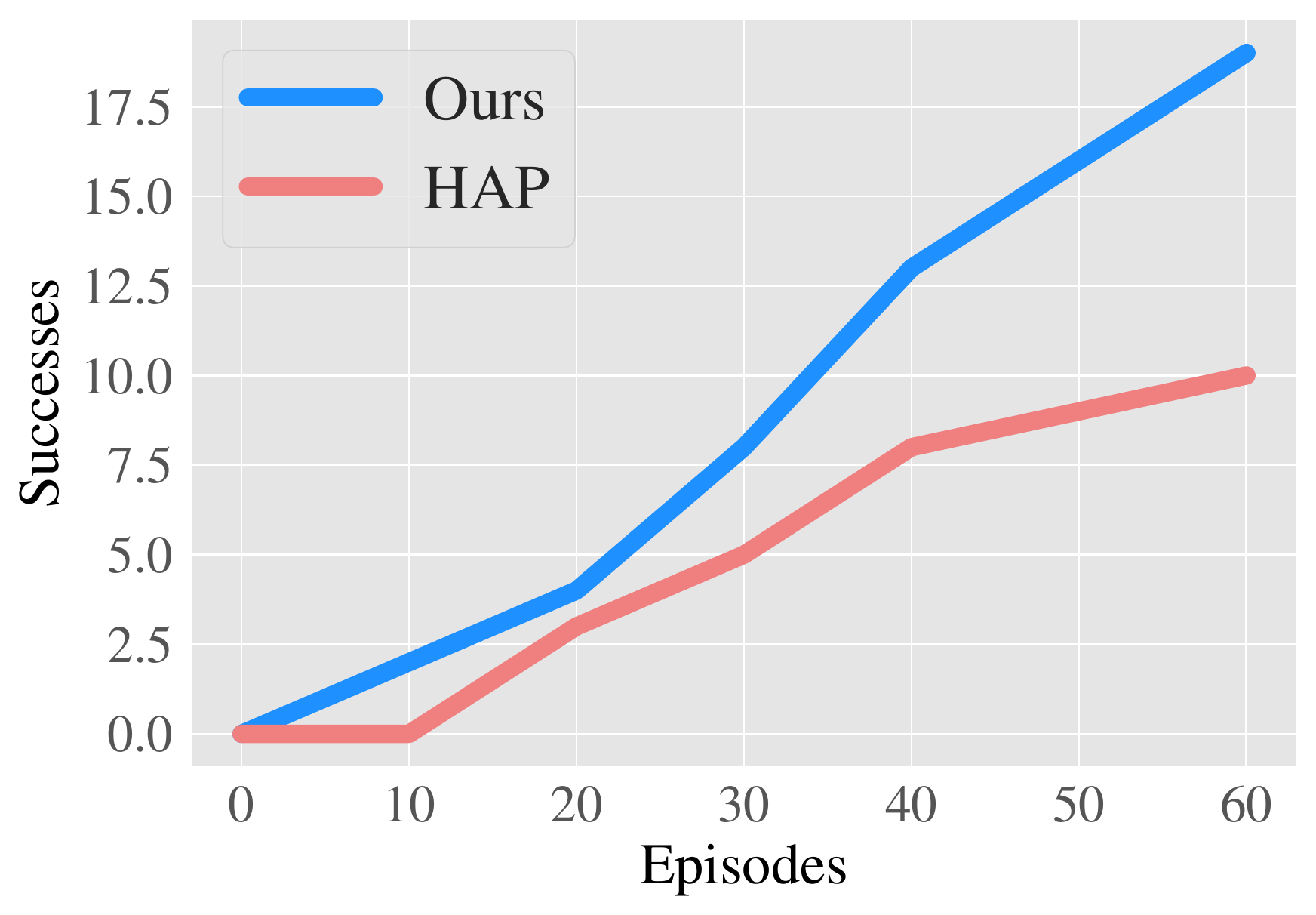}
    \caption{Action Space: Cabinet}
    \end{subfigure}
    \begin{subfigure}[b]{0.32\linewidth}
    \includegraphics[width=\linewidth]{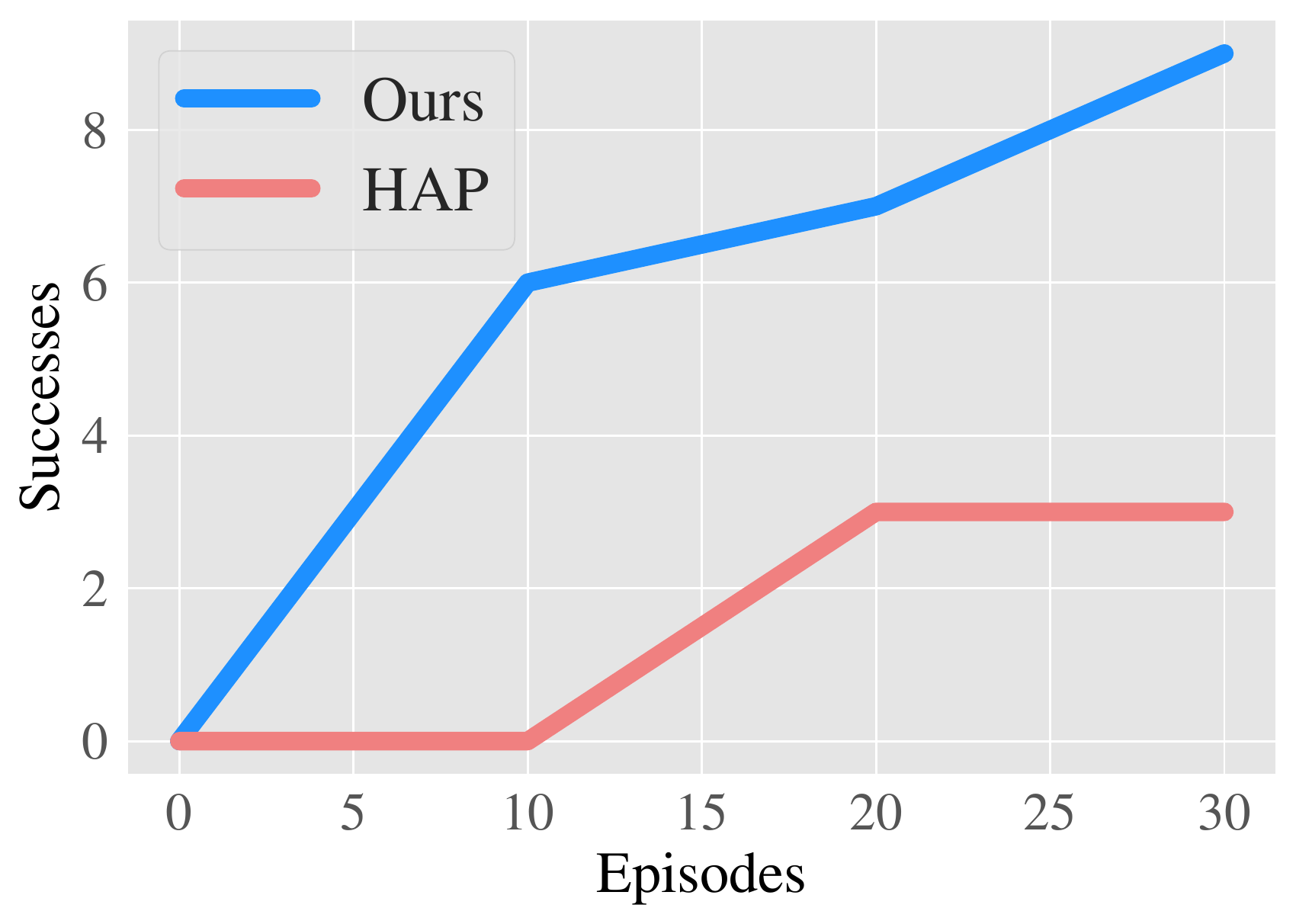}
    \caption{Action Space: Veggies}
    \end{subfigure}
    \caption{\textbf{Exploration} and \textbf{Action Space Parameterization}: Coincidental success (stumbling onto goal configurations) increases multiple folds with \ours in comparison to random exploration or the exploration based on HAP~\cite{hap} in a-d. In e-f, we see the success numbers of using DQN with the discretized action space, for reaching a specified goal image.}
    \label{fig:expl-plots-supp}
\end{figure*}

\noindent{\textbf{Affordance as an Action Space:}} For this learning setup, we parameterize the action space for the robot with the output distribution of our affordance model. We first query the model a large number of times, and then fit Gaussian Mixture Models (GMMs) separately to the $c$ and $\tau$ predictions, with $N_c$ and $N_\tau$ centers respectively. We then define a discrete action space of dimension $N_c$*$N_\tau$, where each action maps to a value in the cross-product space of the centers of the two GMMs. We can now use discrete action-space RL algorithms. We asynchronously sample from the discrete action-space policy, and train it using the RL algorithm. This procedure is described in Algorithm \ref{algo:action_space}. We note that it is important to reset the environment so that images the policy sees are close to the initial image for which the action space was defined. Across experiments we set $N_c$ = $N_\tau$ = 4, $q$ = 2000. For the RL algorithm $RLA$ we use the Deep Q-Network (DQN) \cite{dqn} implementation from the d3rlpy \cite{seno2021d3rlpy} library. We include a visualization of the action space by plotting the ($c,\tau$) values in the cross-product space of the centers of the two GMMs, for VRB and HAP \cite{hap} in Figure \ref{fig:aff_act_space_vis}. We see that for VRB a larger number of the discretized actions are likely to interact with the objects. 

\section{Baselines and Ablations}
\label{sec:exp}

\noindent\textbf{{Baselines}} The baselines we compare to include the approaches from from Liu et al. \cite{hoi} (HOI), Goyal et al. \cite{hap} (HAP) and Natarajan et al., (Hotspots) \cite{hotspots}. In each of these baselines, we used the provided pretrained model. Specficially, for Hotspots \cite{hotspots}, we employ the model trained on EpicKitchens \cite{EPICKITCHENS}, as this is what our approach is also trained on. Similarly, for HAP  \cite{hap} we use the trained model on EpicKitchens also. HOI predicts both a contact point and trajectory, which we execute at test time. The other two approaches predict likely contact regions, from which we sample, as well as a random post contact trajectory. 

\noindent\textbf{{Visual Representation Analysis (Finetuning): }} For the visual representation finetuning experiments we performed in Section 4.5, we use the Imitation Learning Evaluation Framework from R3M~\cite{r3m}, which aims to evaluate the effectiveness of frozen visual representations for performing behavior cloning for robotic control tasks. Following their procedure, we evaluate on three simulated tasks from the Franka Kitchen environment: (1) microwave, (2) slide-door, and (3) door-open. We train the policy using left camera images from their publicly available demonstration dataset, which is collected by an expert state-based reinforcement learning agent and then rendered as image observations. 

For behavior cloning with the R3M encoder, we freeze the pretrained R3M encoder (which uses a ResNet50 base architecture) and finetune a policy on top of it. For behavior cloning with the VRB encoder, we instead use an R3M model which was finetuned for 400 steps with affordance model training as in Section 3.2. Note that this finetuning was performed separately from behavior cloning, and during policy learning our representations are also frozen before being used as input for the downstream policy. For both R3M and VRB, we concatenate the visual embedding and proprioceptive data for input to the downstream policy, and then use a BatchNorm layer followed by a 2-layer MLP to output an action. The downstream policy is trained with a learning rate of 0.001 and a batch size of 32 for 2000 steps.

\noindent\textbf{{Visual Representation Analysis (Feature space distance): }}
For the feature space distance experiments, we compare an R3M model with a VRB model. Both use a ResNet50 base architecture, and the VRB model is obtained by finetuning an R3M model for 100 steps using affordance model training as in Section 3.2. The distances in Figure 8 are computed as the (squared) L2 distances between the features produced by each model for the goal image and current image.

\section{Simulation}

\begin{figure}
    \centering
    \includegraphics[width=0.75\linewidth]{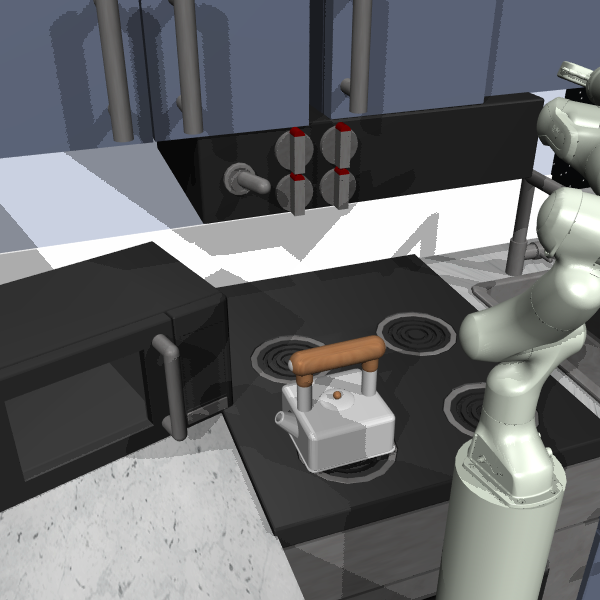}
    \caption{ Simulation Environment from \cite{d4rl}}
\label{fig:sim}
\end{figure}

We also provide a simulation environment benchmark to test our affordances. This is modeled after the Franka-Kitchen environment from the D4RL \cite{d4rl}. In this benchmark, the robot observes images and predicts 3D positions to manipulate, in the exact same way as we deploy the robot in the real world. An image of this environment can be seen in Figure~\ref{fig:sim}. There are three different tasks: turning the light on, opening the microwave and lifting the kettle. These are standard tasks in the D4RL benchmark \cite{d4rl}. We run Paradigm 1 (offline data collection) and provide the success rates for \ours and baselines in Table~\ref{tab:sim}. We can see that \ours significantly outperforms the baselines.

\begin{table}[t]
    \centering
    \resizebox{0.75\linewidth}{!}{%
    \begin{tabular}{c|c|c|c}
        \toprule
        \textbf{Method} & \textbf{Light} & \textbf{Microwave} & \textbf{Kettle}\\
         \midrule
        \textbf{Random} & 0.20 & 0.15 & 0.20\\
        \textbf{HAP} & 0.30 & 0.20 & 0.45\\
        \textbf{HOI} & 0.60 & 0.45 & 0.40 \\
        \textbf{Hotspots} & 0.35 & 0.35 & 0.25 \\
        \midrule
        \textbf{VRB} & \textbf{0.75} & \textbf{0.60} & \textbf{0.55}\\
        \bottomrule
    \end{tabular}
    }
    \vspace{-2mm}
    \caption{\ours on simulation benchmarks.}
    \label{tab:sim}
    \vspace{-4mm}
\end{table}

\section{Codebases}

We use the following codebases: 
\begin{itemize}

\item \href{https://github.com/epic-kitchens/epic-kitchens-100-hand-object-bboxes}{epic-kitchens/epic-kitchens-100-hand-object-bboxes} for extracting detections from 100 DOH \cite{100doh} for EpicKitchens \cite{EPICKITCHENS}.  

\item \href{https://github.com/stevenlsw/hoi-forecast}{stevenlsw/hoi-forecast} for Skin segmentation code and HOI baseline \cite{hoi}. 

\item \href{https://github.com/uiuc-robovision/hands-as-probes}{uiuc-robovision/hands-as-probes} for HAP baseline \cite{hap}.  

\item \href{https://github.com/Tushar-N/interaction-hotspots}{Tushar-N/interaction-hotspots} for Hotspots baseline \cite{hotspots}. 

\item \href{https://github.com/facebookresearch/r3m}{facebookresearch/r3m} for R3M visual features \cite{r3m}. 

\item \href{https://github.com/wkentaro/labelme}{wkentaro/labelme} for getting masks for robot and  \item \href{https://pytorch.org/tutorials/intermediate/torchvision_tutorial.html}{Torchvision tutorial} for a Mask-RCNN \cite{he2017mask} implementation. 

\item \href{https://github.com/takuseno/d3rlpy}{takuseno/d3rlpy} \cite{seno2021d3rlpy} for DQN \cite{dqn} implementation. 

\item \href{https://github.com/facebookresearch/fairo/tree/main/polymetis}{facebookresearch/polymetis} \cite{Polymetis2021} as the base for the controller for the Franka Arm.
\end{itemize}

\end{document}